\documentclass{article}

\usepackage{microtype}
\usepackage{graphicx}
\usepackage{subfigure}
\usepackage{booktabs}
\usepackage{multirow}
\usepackage{float}
\usepackage{array}
\usepackage{amsmath,amssymb}
\usepackage{xcolor}
\usepackage{colortbl}
\definecolor{oursrow}{RGB}{230,239,251}  %
\usepackage{tikz}
\usetikzlibrary{positioning,arrows.meta,calc}
\usepackage{hyperref}
\usepackage{xspace}

\def\showauthors{1}
\ifdefined\showauthors\usepackage[accepted]{mlsys2025}\else\usepackage{mlsys2025}\fi

\newcommand{\sys}{\textsc{Moira}\xspace}  %
\newcommand{\kest}{\textsf{Estimate}\xspace}
\newcommand{\ksel}{\textsf{Select}\xspace}
\newcommand{\katt}{\textsf{Attend}\xspace}
\newcommand{\kcmb}{\textsf{Merge}\xspace}
\newcommand{\mvar}{\textsc{mvar}\xspace}
\newcommand{\gam}{\gamma}

\usepackage{iftex}
  \usepackage[T1]{fontenc}
  \usepackage{tgtermes}
  \usepackage[scale=0.86]{tgheros}

\mlsystitlerunning{MOIRA: Mass-Oriented Indexing with Ragged Attention}

\begin{document}

\twocolumn[
\mlsystitle{MOIRA: Mass-Oriented Indexing with Ragged Attention\\ for Long-Context Decoding}

\begin{mlsysauthorlist}
\mlsysauthor{Dich Nhat Minh Nguyen}{}  %
\mlsysauthor{Tran Dang Duong Nguyen}{}  %
\end{mlsysauthorlist}

\mlsyskeywords{sparse attention, LLM inference, decode, KV cache, GPU kernels}

\vskip 0.3in

\begin{abstract}
Long-context decoding is limited by memory bandwidth, because every output token reads the KV cache of every layer.
Sparse decoding reduces this cost by reading only part of the KV cache. We observe that the number of pages a query
needs varies widely across KV heads, layers and steps. Fixed budgets are simple, but they are sized for demanding cases
and tuned per workload; adaptive budgets follow this variation more flexibly, but existing designs pay for it with extra
selection cost or training. At the kernel level, FlashAttention-3 (FA3) and FlashInfer are designed for rows of similar length: with page
lists whose length differs per KV head, they pad the lists, which forfeits much of the sparse saving, leave thread blocks unbalanced,
or rely on a host-side plan that runs outside the CUDA graph.
We propose \sys, a training-free sparse decode path in vLLM whose budget adapts per KV head and per layer. For every
request, layer, KV head and step, a coverage rule keeps the smallest set of pages whose estimated attention mass reaches
a fraction $\gam$. A new kernel, self-planning attention, lets each thread block derive its own share of the work from
the list lengths, so the whole decode step stays inside the CUDA graph.
On an H200, at RULER's 128k context, \sys with $\gam{=}0.99$ matches dense accuracy while reading about 30\% of the
pages and reduces the time per output token (TPOT) by 2.2--2.5$\times$ relative to dense FA3; with $\gam{=}0.98$ it reduces TPOT by 2.7$\times$ and
stays within the noise of dense. Under high serving load it raises throughput by up to 51\%. These results suggest that a budget
adapted per head and layer, paired with a kernel that keeps such budgets inside the CUDA graph, makes sparse decoding
both flexible and fast.
\end{abstract}

]

\section{Introduction}
\label{sec:intro}

Long-context decoding is limited by memory bandwidth: every output token reads the KV cache of every layer. Existing
work reduces this cost by reading only part of the KV cache at each decode step, and falls into two groups.
\emph{Fixed-budget methods}, such as Quest~\citep{tang2024quest} and the trained sparse attention of recent production
models~\citep{deepseek2025v32,lai2026msa,qiu2026qwen38next}, read the same number of tokens, the \emph{budget}, for
every head and every step.
\emph{Adaptive-budget methods} let the budget vary with the query: Twilight~\citep{lin2025twilight} selects tokens per
query head with top-$p$, FFD~\citep{liu2026ffd} keeps the blocks whose score, estimated from a low-bit copy of the keys,
exceeds a threshold, and Elastic Threshold Attention~\citep{haris2026eta} uses a learned threshold.

However, the budget that is actually needed varies widely across KV heads, across layers and across steps
(Figure~\ref{fig:motivation}). A fixed budget cannot follow this variation: it must be sized for the hardest case, so
most heads over-read, and it must be re-tuned for each workload. Adaptive methods already let the
budget vary per head, layer and step, but they pay for it elsewhere: Twilight selects per query head and plans its
lists on the host, FFD scans a low-bit copy of every key at every step, and Elastic Threshold Attention requires
training.
On the kernel side, the selected pages are read by decode kernels such as FlashAttention-3 (FA3)~\citep{shah2024flashattention3} and
FlashInfer~\citep{ye2025flashinfer}, which are built for rows of similar length. When each KV head has a list of its own
length, they either pad the lists, which forfeits much of the sparse saving, read the exact lengths but leave thread
blocks unbalanced, or balance them with a host-side plan that must run outside the CUDA graph before every step.

In this paper, we propose \textbf{\sys} (\emph{Mass-Oriented Indexing with Ragged Attention}), a
training-free sparse decode path inside vLLM~\citep{kwon2023vllm} that addresses both limitations. First, \sys applies a
\emph{coverage rule}: for every request, layer, KV head and step, it keeps the smallest set of pages whose estimated
attention mass reaches a fraction $\gam$, and one fused kernel evaluates the rule. Second, a new kernel,
\emph{self-planning attention}, lets each thread block derive its own share of the work from the list lengths in device
memory, so the whole decode step stays in vLLM's CUDA graphs. On an H200 at 128k, \sys with $\gam{=}0.99$ matches
dense accuracy on RULER~\citep{hsieh2024ruler} with \textbf{\mbox{27--33}\%} of the page reads and reduces the
time per output token (TPOT) by \textbf{2.2--2.5$\times$} relative to dense FA3.

\vspace*{\fill}\newpage
Our contributions are:
{\setlength\topsep{2pt}\setlength\partopsep{0pt}
\begin{itemize}\setlength\itemsep{2pt}\setlength\parsep{0pt}\setlength\parskip{0pt}
\item \textbf{A coverage rule for per-head, per-layer budgets.} We show that the number of pages a query needs varies
across KV heads, layers and steps, and that one coverage rule follows this variation across models, context lengths
and tasks with $\gam$ fixed across them, whereas a fixed budget must be re-tuned for each case or profiled per KV head on the target model (Sections~\ref{sec:motivation},
\ref{sec:eval-acc} and~\ref{sec:eval-ablation}). This rule is our main contribution.
\item \textbf{A kernel that keeps such budgets in the CUDA graph.} Existing decode kernels pad lists of different
lengths, leave thread blocks unbalanced, or balance them with a host-side plan. Self-planning attention balances them on the GPU, so the rule runs
inside vLLM's CUDA graphs without padding or a host plan (Figure~\ref{fig:kern}, Section~\ref{sec:design}).
\item \textbf{System and evaluation.} We implement both in vLLM and compare \sys with FA3, FlashInfer, Quest, Twilight and
FFD. \sys lowers TPOT by 1.38--2.69$\times$ on synthetic batches of 8k--130k tokens, raises serving throughput by
45--51\% under high load, and matches dense accuracy on long reasoning (AIME) and on RULER at 128k (Section~\ref{sec:eval}).
\end{itemize}}

\section{Background and Motivation}
\label{sec:background}

\subsection{Background}
\label{sec:bg}

\subsubsection{Decode attention in a paged serving engine}
Each decode step generates one token per request, whose query attends over the request's whole KV cache. With
grouped-query attention (GQA), $G$ query heads share one KV head, so the unit of work is a \emph{row}: one (request,
KV head) pair whose keys and values are read once for all $G$ query heads. Four requests on Llama-3.1-8B (8 KV heads)
give 32 rows per step. vLLM~\citep{kwon2023vllm} stores the KV cache in fixed-size \emph{pages} (64 tokens here);
each request's pages are listed in a block table, through which the kernel reads its rows.

These 32 rows are far fewer than the 132 streaming multiprocessors (SMs) of an H200. To fill the GPU, decode kernels cut long rows into chunks, one per
thread block, and merge the partial results through their log-sum-exp (\emph{split-KV}, or
Flash-Decoding~\citep{dao2023flashdecoding,hong2024flashdecodingpp}). The cut requires a \emph{plan}: which chunk of
which row each thread block processes. FA3~\citep{shah2024flashattention3} builds this plan on the GPU from the shape of
the batch; FlashInfer~\citep{ye2025flashinfer} builds it on the host, in \texttt{plan()}, from the length of each
request.

A decode step lasts a few milliseconds, so host launch cost matters; vLLM captures the step once as a \emph{CUDA
graph} and replays it at every step, which re-launches the captured kernels without host involvement. Host work that depends on the
current step, such as FlashInfer's \texttt{plan()}, cannot be part of the graph: it must run before every replay, or be
removed.

\subsubsection{Query-aware sparse decode}
Query-aware methods estimate the attention the current query puts on each page and read only the pages that
matter. Let $a_P$ be the fraction of a row's attention mass that falls on page $P$, averaged over the $G$ query heads.
The smallest set of pages that covers a fraction $\gam$ of the mass is the \emph{oracle coverage set}
\begin{equation}
\mathcal{C}^\star(\gam) = \arg\min_{\mathcal{C}} |\mathcal{C}| \quad \text{s.t.} \quad
\textstyle\sum_{P\in\mathcal{C}} a_P \ge \gam ,
\label{eq:oracle}
\end{equation}
and its size $|\mathcal{C}^\star|$ is the \emph{budget} the row actually needs. Methods differ in how they approximate
it. Quest~\citep{tang2024quest} bounds each page's largest logit with a per-channel min/max box of its keys and keeps a
fixed number of tokens. Twilight~\citep{lin2025twilight} further prunes Quest's candidates separately for each query
head with a \emph{top-$p$} rule: it estimates the attention weights of the candidates from int4 keys and keeps the
fewest tokens whose total weight reaches a fraction $p$ (here $p{=}0.85$), so concentrated attention keeps fewer
tokens. FFD~\citep{liu2026ffd} keeps every block whose largest logit, estimated from a 2-bit copy of the keys, lies within
$\delta$ of a maximum estimated from the sink and local tokens. Quest and Twilight then read exact keys and values for the
kept pages; FFD rebuilds the kept keys from the 2-bit copy plus an 8-bit residual and reads exact values.

\begin{figure*}[t]
  \centering
  \includegraphics[width=\textwidth]{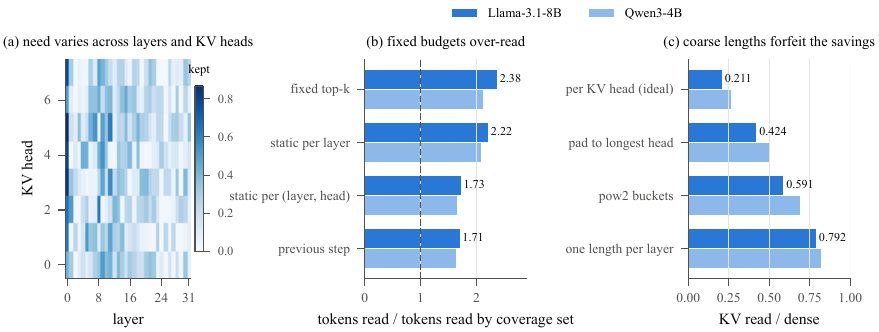}
  \vspace{-6mm}
  \caption{\textbf{The budget a row needs varies, and fixed budgets or coarse lengths forfeit the savings.}
  Oracle coverage sets $\mathcal{C}^\star(0.95)$ over 60 LongBench requests, 64-token pages.
  (a)~Mean size of the set, as a fraction of the context, for each (layer, KV head) of Llama-3.1-8B: the first layer
  needs most of its context, while most layers need only a tenth.
  (b)~Tokens read by budget policies relative to the coverage set, when each policy must give 90\% of the
  rows, over all layers and steps, at least 0.95 of the mass: every fixed or profiled budget reads 1.7--2.4$\times$ as many
  tokens as the coverage set (dashed line).
  (c)~Fraction of the dense KV cache read by batches of 32 requests (total over decode steps 16--256) when the
  per-(layer, KV head) lengths are coarsened: padding to the longest
  head doubles the reads.}
  \label{fig:motivation}
\end{figure*}

\subsection{Motivation}
\label{sec:mot}

\subsubsection{The budget must adapt per KV head and per layer}
\label{sec:motivation}
\paragraph{The needed budget varies across layers, KV heads and steps.}
We record $\mathcal{C}^\star(0.95)$ at every decode step of 60 LongBench requests on Llama-3.1-8B and Qwen3-4B, for
every layer and KV head (Figure~\ref{fig:motivation}).
On Llama-3.1-8B a row needs 21.6\% of the context on average to keep 95\% of the mass: 68\% in the first layer but
only 9\% in several others (Figure~\ref{fig:motivation}a). The coefficient of variation (CV, standard deviation over
mean) of the page count is 0.61 across layers, 0.74 across the KV heads of one layer at one step, and 0.53 across the
steps of one (layer, KV head). Totals over all layers vary much less (CV 0.13--0.18 across KV heads, requests and
steps): the variation appears at the granularity of (layer, KV head), which is where the budget must be set.

\paragraph{Fixed budgets over-read.}
A fixed budget must cover the rows that need the most. With the smallest budget that lets 90\% of the rows keep 95\%
of the mass, a single token count shared by all rows reads 2.38$\times$ the coverage set (Figure~\ref{fig:motivation}b). Even a budget profiled offline for every (layer, KV head)
reads 1.73$\times$, because within one head the need still changes with the query. These counts use the exact mass (an oracle
analysis); because a real estimator keeps more pages than the oracle set, Section~\ref{sec:eval-ablation} compares the
two end to end at equal reads.

\paragraph{One set per KV head is the right granularity.}
If every KV head of a layer reads as many pages as its longest head, page reads double, from 21\% to 42\% of the dense
cache over a batch of 32 requests (Figure~\ref{fig:motivation}c). Selecting per query head, as Twilight does, needs 1.51$\times$ the pages at the
same output error, because the $G$ query heads share one KV head and every page any of them selects must be read.

\emph{The budget should therefore be chosen per (request, layer, KV head, step), from the attention mass averaged over
the query heads of the group.} This is the coverage rule of Section~\ref{sec:select}.

\begin{figure*}[t]
  \centering
  \begin{tikzpicture}[x=0.33cm, y=0.30cm, font=\scriptsize,
  wa/.style={fill=blue!55, draw=white, line width=0.5pt},
  wb/.style={fill=blue!28, draw=white, line width=0.5pt},
  pad/.style={fill=red!22, draw=white, line width=0.5pt},
  lab/.style={align=center, text width=3.3cm},
  tim/.style={align=center, font=\footnotesize},
  bad/.style={text=red!70!black},
]
\newcommand{\wk}[5]{\path[#5] ({#1+#3},{-1.2*#2}) rectangle ++(#4,-1.0);}

\begin{scope}
  \node[lab, anchor=south] at (4,0.3) {\textbf{(a) FA3, padded to max}};
  \wk{0}{0}{0}{4}{wa}\wk{0}{0}{4}{4}{wb}
  \wk{0}{1}{0}{3}{wa}\wk{0}{1}{3}{1}{pad}\wk{0}{1}{4}{4}{pad}
  \wk{0}{2}{0}{4}{wa}\wk{0}{2}{4}{1}{wb}\wk{0}{2}{5}{3}{pad}
  \wk{0}{3}{0}{2}{wa}\wk{0}{3}{2}{2}{pad}\wk{0}{3}{4}{4}{pad}
  \node[tim] at (4,-5.6) {\textbf{492\,$\mu$s} (dense 489)};
  \node[lab] at (4,-6.75) {\textcolor{red!70!black}{reads unselected pages}};
\end{scope}

\begin{scope}[xshift=3.45cm]
  \node[lab, anchor=south] at (4,0.3) {\textbf{(b) FA3, pow-2 buckets}};
  \wk{0}{0}{0}{4}{wa}\wk{0}{0}{4}{4}{wb}
  \wk{0}{1}{0}{3}{wa}\wk{0}{1}{3}{1}{pad}
  \wk{0}{2}{0}{4}{wa}\wk{0}{2}{4}{1}{wb}\wk{0}{2}{5}{3}{pad}
  \wk{0}{3}{0}{2}{wa}
  \node[tim] at (4,-5.6) {\textbf{281\,$\mu$s}};
  \node[lab] at (4,-6.75) {\textcolor{red!70!black}{still reads padding}};
\end{scope}

\begin{scope}[xshift=6.9cm]
  \node[lab, anchor=south] at (4,0.3) {\textbf{(c) FA3 varlen}};
  \wk{0}{0}{0}{4}{wa}\wk{0}{0}{4}{4}{wb}
  \wk{0}{1}{0}{1.5}{wa}\wk{0}{1}{1.5}{1.5}{wb}
  \wk{0}{2}{0}{2.5}{wa}\wk{0}{2}{2.5}{2.5}{wb}
  \wk{0}{3}{0}{1}{wa}\wk{0}{3}{1}{1}{wb}
  \node[tim] at (4,-5.6) {\textbf{238\,$\mu$s}};
  \node[lab] at (4,-6.75) {\textcolor{red!70!black}{exact, but unbalanced}};
\end{scope}

\begin{scope}[xshift=10.35cm]
  \node[lab, anchor=south] at (4,0.3) {\textbf{(d) FlashInfer}};
  \wk{0}{0}{0}{2}{wa}\wk{0}{0}{2}{2}{wb}\wk{0}{0}{4}{2}{wa}\wk{0}{0}{6}{2}{wb}
  \wk{0}{1}{0}{2}{wa}\wk{0}{1}{2}{1}{wb}
  \wk{0}{2}{0}{2}{wa}\wk{0}{2}{2}{2}{wb}\wk{0}{2}{4}{1}{wa}
  \wk{0}{3}{0}{2}{wa}
  \node[tim] at (4,-5.6) {\textbf{194\,$\mu$s} + \textbf{464\,$\mu$s} host plan};
  \node[lab] at (4,-6.75) {\textcolor{red!70!black}{host plan, not in CUDA graph}};
\end{scope}

\begin{scope}[xshift=13.8cm]
  \node[lab, anchor=south] at (4,0.3) {\textbf{(e) Self-planning (ours)}};
  \wk{0}{0}{0}{2}{wa}\wk{0}{0}{2}{2}{wb}\wk{0}{0}{4}{2}{wa}\wk{0}{0}{6}{2}{wb}
  \wk{0}{1}{0}{2}{wa}\wk{0}{1}{2}{1}{wb}
  \wk{0}{2}{0}{2}{wa}\wk{0}{2}{2}{2}{wb}\wk{0}{2}{4}{1}{wa}
  \wk{0}{3}{0}{2}{wa}
  \node[tim] at (4,-5.6) {\textbf{201\,$\mu$s}};
  \node[lab] at (4,-6.75) {\textcolor{green!40!black}{exact, balanced, in graph}};
\end{scope}
\end{tikzpicture}
  \vspace{-4mm}
  \caption{\textbf{Existing decode kernels on per-head lists of different lengths.} Each bar is one (request, KV head)
  row with its selected pages; each segment is the work of one thread block; red is padding. Times: attention of one
  layer of Llama-3.1-8B on lists from real decode steps, scaled to 32k context $\times$ 16 requests (H200). Only self-planning
  attention reads exactly the selected pages, gives every thread block a bounded, nearly equal amount of work, and runs inside a CUDA
  graph.}
  \label{fig:kern}
\end{figure*}
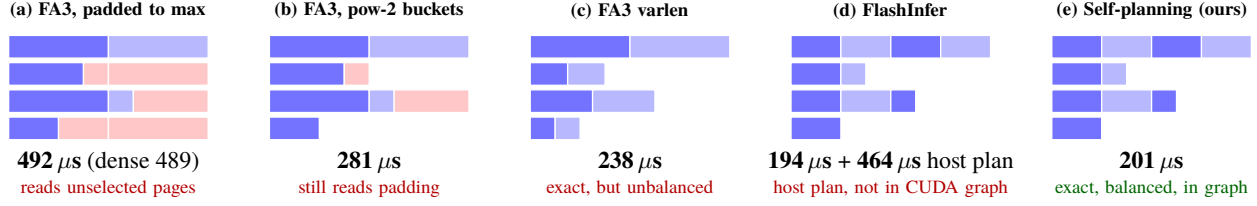

\subsubsection{Existing kernels forfeit the savings}
\label{sec:motivation-kernel}
With an adaptive per-head budget, each row's page count is known only on the GPU, within the step, after selection.
A suitable kernel must (i)~read only the selected pages, (ii)~divide the work evenly
among thread blocks, and (iii)~need no host involvement within the step, so that it can run inside a CUDA graph.
Figure~\ref{fig:kern} shows that every existing kernel misses at least one condition.

\paragraph{Padding reads pages that were not selected.}
A rectangular grid gives every row enough thread blocks for the longest row and pads the rest; the padding is read
like real pages, so sparse attention is as slow as dense (Figure~\ref{fig:kern}a). Power-of-two buckets reduce the
padding but remain 1.4$\times$ slower than reading exactly the selected pages (Figure~\ref{fig:kern}b). Both miss~(i).

\paragraph{Exact lengths are not enough.}
FA3's varlen mode takes a length per row and reads only the selected pages, but its split scheduler does not balance
rows of different lengths: thread blocks of long rows finish last while the others sit idle (Figure~\ref{fig:kern}c).
At equal total work, FA3 slows by 23--32\% as row lengths spread. FA3 misses~(ii).

\paragraph{Balance planned on the host does not fit the step.}
FlashInfer cuts every row into equal chunks, balancing the load (Figure~\ref{fig:kern}d), but computes this plan on
the host from the row lengths. With adaptive lists the lengths change at every layer and step, so the host must fetch
them from the GPU and re-plan each time; at 32k$\times$16 the plan alone costs 2.4$\times$ the kernel time and cannot
be part of a CUDA graph. FlashInfer misses~(iii).

\paragraph{Planning from past steps overflows.}
One could instead reserve thread-block slots per row at capture time, sized from earlier steps. But lengths change
between steps: with slots sized from the previous step, 40\% of the rows overflow, and a safety margin that cuts the
overflow to 2.4\% reserves 1.77$\times$ the needed slots, so the kernel again does extra work.

\emph{The kernel must therefore partition the work itself, on the GPU, within the step, from the lengths the selection
has just produced.} Self-planning attention (Section~\ref{sec:balanced}) does exactly this and meets all three
conditions (Figure~\ref{fig:kern}e).

\section{Design}
\label{sec:design}

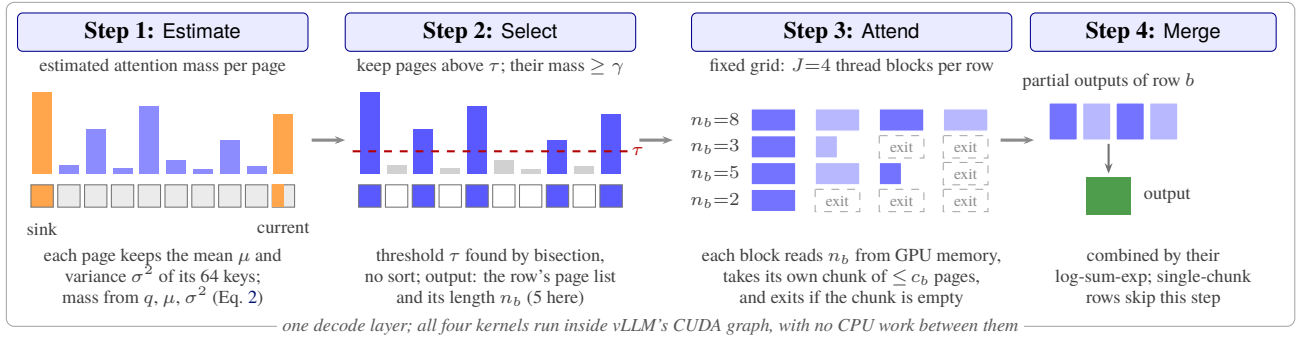
\begin{figure*}[t]
  \centering
  \resizebox{\textwidth}{!}{\begin{tikzpicture}[x=1cm, y=1cm, font=\scriptsize, >={Stealth[length=4pt]},
  head/.style={draw=blue!55!black, fill=blue!8, rounded corners=2pt, minimum width=3.9cm, minimum height=0.5cm,
               align=center, font=\footnotesize},
  pg/.style={draw=black!55, fill=black!8, minimum size=0.3cm, inner sep=0pt},
  ex/.style={fill=orange!70},
  est/.style={fill=blue!45},
  keep/.style={fill=blue!65},
  drop/.style={fill=black!18},
  ca/.style={fill=blue!55, draw=white, line width=0.5pt},
  cb/.style={fill=blue!28, draw=white, line width=0.5pt},
  idle/.style={draw=black!35, dashed},
  note/.style={align=center, text=black!75},
  flowarrow/.style={->, very thick, draw=black!45},
]
\newcommand{\pbar}[3]{\path[#3] (#1-0.13,0.55) rectangle (#1+0.13,0.55+#2);}
\def\hs{1.1,0.12,0.6,0.08,0.9,0.18,0.06,0.45,0.1,0.8}

\begin{scope}
  \node[head] at (2,2.45) {\textbf{Step 1:} \kest};
  \foreach \h [count=\i from 0] in \hs {
    \pgfmathsetmacro{\x}{0.38+0.36*\i}
    \ifnum\i=0 \node[pg, ex] at (\x,0.25) {}; \pbar{\x}{\h}{ex}
    \else\ifnum\i=9 \node[pg] at (\x,0.25) {}; \path[ex] (\x-0.15,0.1) rectangle (\x,0.4); \pbar{\x}{\h}{ex}
    \else \node[pg] at (\x,0.25) {}; \pbar{\x}{\h}{est}\fi\fi
  }
  \node[note, anchor=south] at (2,1.75) {estimated attention mass per page};
  \node[note, anchor=north] at (0.38,0.02) {sink};
  \node[note, anchor=north] at (3.62,0.02) {current};
  \node[note] at (2,-0.85) {each page keeps the mean $\mu$ and\\variance $\sigma^2$ of its 64 keys;\\mass from $q$, $\mu$, $\sigma^2$ (Eq.~\ref{eq:mvar})};
\end{scope}

\begin{scope}[xshift=4.4cm]
  \node[head] at (2,2.45) {\textbf{Step 2:} \ksel};
  \foreach \h [count=\i from 0] in \hs {
    \pgfmathsetmacro{\x}{0.38+0.36*\i}
    \pgfmathparse{(\h>0.3) ? 1 : 0}
    \ifnum\pgfmathresult=1 \node[pg, fill=blue!65] at (\x,0.25) {}; \pbar{\x}{\h}{keep}
    \else \node[pg, fill=white] at (\x,0.25) {}; \pbar{\x}{\h}{drop}\fi
  }
  \draw[red!70!black, dashed, thick] (0.15,0.85) -- (3.85,0.85) node[right, inner sep=1pt] {$\tau$};
  \node[note, anchor=south] at (2,1.75) {keep pages above $\tau$; their mass $\ge\gam$};
  \node[note] at (2,-0.85) {threshold $\tau$ found by bisection,\\no sort; output: the row's page list\\and its length $n_b$ (5 here)};
\end{scope}

\begin{scope}[xshift=8.8cm]
  \node[head, minimum width=4.3cm] at (2.45,2.45) {\textbf{Step 3:} \katt};
  \node[note, anchor=south] at (2.45,1.75) {fixed grid: $J{=}4$ thread blocks per row};
  \newcommand{\blk}[4]{\path[#4] ({1.1+0.86*#2},{1.42-0.36*#1}) rectangle ++({0.3*#3},-0.3);}
  \newcommand{\idl}[2]{\draw[idle] ({1.1+0.86*#2},{1.42-0.36*#1}) rectangle ++(0.6,-0.3);
                       \node[text=black!45, font=\tiny] at ({1.4+0.86*#2},{1.27-0.36*#1}) {exit};}
  \foreach \r/\n in {0/8, 1/3, 2/5, 3/2} {\node[note, anchor=east] at (1.07,{1.27-0.36*\r}) {$n_b{=}\n$};}
  \blk{0}{0}{2}{ca}\blk{0}{1}{2}{cb}\blk{0}{2}{2}{ca}\blk{0}{3}{2}{cb}
  \blk{1}{0}{2}{ca}\blk{1}{1}{1}{cb}\idl{1}{2}\idl{1}{3}
  \blk{2}{0}{2}{ca}\blk{2}{1}{2}{cb}\blk{2}{2}{1}{ca}\idl{2}{3}
  \blk{3}{0}{2}{ca}\idl{3}{1}\idl{3}{2}\idl{3}{3}
  \node[note] at (2.45,-0.85) {each block reads $n_b$ from GPU memory,\\takes its own chunk of $\le$\,$c_b$ pages,\\and exits if the chunk is empty};
\end{scope}

\begin{scope}[xshift=13.65cm]
  \node[head, minimum width=3.3cm] at (1.65,2.45) {\textbf{Step 4:} \kcmb};
  \foreach \j/\s in {0/ca, 1/cb, 2/ca, 3/cb} {\path[\s] ({0.25+0.45*\j},1.0) rectangle ++(0.38,0.5);}
  \node[note, anchor=south] at (1.05,1.55) {partial outputs of row $b$};
  \draw[->, thick, black!55] (1.05,0.95) -- (1.05,0.55);
  \path[fill=green!45!black!70] (0.75,0.0) rectangle ++(0.6,0.5);
  \node[note, anchor=west] at (1.4,0.25) {output};
  \node[note] at (1.65,-0.85) {combined by their\\log-sum-exp; single-chunk\\rows skip this step};
\end{scope}

\draw[flowarrow] (4.0,1.0) -- (4.45,1.0);
\draw[flowarrow] (8.4,1.0) -- (8.85,1.0);
\draw[flowarrow] (13.25,1.0) -- (13.7,1.0);
\draw[black!35, rounded corners=3pt] (-0.1,-1.5) rectangle (17.2,2.85);
\node[fill=white, inner sep=2pt, font=\scriptsize\itshape, text=black!65] at (8.55,-1.5)
  {one decode layer; all four kernels run inside vLLM's CUDA graph, with no CPU work between them};
\end{tikzpicture}}
  \vspace{-5mm}
  \caption{\textbf{One decode layer of \sys, for one (request, KV head) row.} Step~1 approximates the attention mass of
  each page; Step~2 keeps the fewest pages whose total mass reaches $\gam$; Step~3 reads those pages, with each thread
  block taking one chunk of the row by itself; Step~4 combines the chunks. Values are illustrative; Step~3 uses the row
  lengths of Figure~\ref{fig:kern}.}
  \label{fig:overview}
\end{figure*}

\sys replaces vLLM's FA3 decode attention with four kernels that run in sequence at every layer
(Figure~\ref{fig:overview}), each over all rows $b$ of the step:
\textbf{Step~1} (\kest) approximates the attention mass of each page from a small summary, without reading the keys
(Section~\ref{sec:mvar});
\textbf{Step~2} (\ksel) keeps the fewest pages that cover a fraction $\gam$ of the mass and writes each row's page list
(Section~\ref{sec:select});
\textbf{Step~3} (\katt) reads exactly the pages in the list, with the work spread evenly over the thread blocks
(Section~\ref{sec:balanced});
\textbf{Step~4} (\kcmb) combines the partial results of each row.
\ksel chooses a budget per request, layer, KV head and step; \katt partitions the work on the GPU without the host;
\kest keeps the cost of selection small relative to the reads it saves.

\subsection{Page summaries and the \mvar estimate}
\label{sec:mvar}
A page holds $S{=}64$ tokens. For each page $P$ and each KV head, \sys stores a \emph{page summary}: the per-channel
mean $\mu_P$ and variance $\sigma_P^2$ of the 64 keys ($2D$ values, written once when the page fills). From this
summary, \kest approximates the attention mass of the page for query head $g$, $m_{g,P}=\sum_{k\in P} e^{s\,q_g\cdot k}$
with $s=1/\sqrt{D}$ the softmax scale, without reading the keys:
\begin{equation}
\log \hat m_{g,P} = \underbrace{s\, q_g\!\cdot\!\mu_P}_{\text{mean logit}} \;+\; \underbrace{\tfrac{1}{2}\,(s\, q_g)^2\!\cdot\!\sigma_P^2}_{\text{spread of the logits}} \;+\; \log S .
\label{eq:mvar}
\end{equation}
We call this estimate \mvar (mean and variance). Eq.~\ref{eq:mvar} is the logarithm of $S$ times the expectation of
$e^{s\,q_g\cdot k}$ when the keys of a page are treated as independent Gaussians per channel, the same second-order approximation that CompKV~\citep{huang2026compkv} uses. The second term matters
because a sum of exponentials is dominated by its largest logits, which the mean hides: on real pages that
contain a few outlying keys, the mean term alone underestimates the mass by a factor of $e^{5}$--$e^{7}$ and the page is
dropped, whereas \mvar underestimates it by a factor of about $e^{3}$ (Appendix~\ref{app:design}). Unlike Quest~\citep{tang2024quest}, which bounds the \emph{largest} logit, \mvar estimates
the \emph{total} mass, which is exactly the quantity the coverage rule accumulates. The first page (the attention
sink~\citep{xiao2024streamingllm}) and the current page, which has no summary yet, are computed exactly.

\subsection{The coverage rule}
\label{sec:select}
\ksel averages the estimated masses over the $G$ query heads of the group (after a softmax per head), giving each
page's attention share $\bar\pi_P$, and keeps pages in decreasing order of share until their sum reaches $\gam$. For example, with four pages of $\bar\pi$ = 0.70, 0.20, 0.08 and 0.02, $\gam{=}0.9$ keeps two pages
and $\gam{=}0.98$ keeps three: the number of kept pages is an \emph{output} of the rule, not a preset parameter. The
sink $P_0$ and the current page $P_{\text{cur}}$ are always kept. Written as a threshold:
\begin{equation}
\begin{aligned}
\mathcal{K}_b &= \{P_0, P_{\text{cur}}\} \cup \{P \in \mathcal{P}_b : \bar\pi_P \ge \tau_b\},\\
\tau_b &= \max\{\tau : \textstyle\sum_{P\in\mathcal{P}_b,\,\bar\pi_P\ge\tau}\bar\pi_P \ge \gam - \bar\pi_{P_0} - \bar\pi_{P_{\text{cur}}}\},
\end{aligned}
\label{eq:topp}
\end{equation}
where $\mathcal{P}_b$ holds the row's pages other than $P_0$ and $P_{\text{cur}}$, so the two forced pages are counted
once. This is top-$p$ over pages, the estimated counterpart of the oracle coverage set (Eq.~\ref{eq:oracle}). $\gam$ is the
only parameter (default 0.98 for every head). Averaging over the $G$ query heads yields one list per group, so each
page is read once for all $G$ heads (Section~\ref{sec:motivation}).

\begin{algorithm}[t]
  \caption{Coverage rule for one row $b$ (\ksel)}
  \label{alg:select}
  \small
\begin{algorithmic}[1]
  \STATE {\bfseries Input:} log-masses $\ell_{g,P}$ for $g\le G$ and pages $P_0,\dots,P_{\text{cur}}$; coverage $\gam$;
  $K_B$ thresholds per pass
  \STATE {\bfseries Output:} page list $L_b$ and its length $n_b$
  \STATE $\bar\pi \gets \frac1G\sum_g \mathrm{softmax}(\ell_{g,\cdot})$ \hfill $\triangleright$ group-mean mass
  \STATE $t \gets \gam - \bar\pi_{P_0} - \bar\pi_{P_{\text{cur}}}$;\ \ $lo \gets 0$;\ \ $hi \gets \max_{P\in\mathcal{P}_b} \bar\pi_P$
  \FOR{$24/\log_2 K_B$ passes}
    \STATE $\tau_k \gets lo + (hi - lo)\,k/K_B$ for $k < K_B$ \hfill $\triangleright$ one reduction
    \STATE $M_k \gets \sum_{P\in\mathcal{P}_b,\,\bar\pi_P \ge \tau_k} \bar\pi_P$
    \STATE $lo \gets \max\{\tau_k : M_k \ge t\}$
    \STATE $hi \gets \min(\{\tau_k : M_k < t\}\cup\{hi\})$
  \ENDFOR
  \STATE $L_b \gets [P_0] \,\Vert\, [P\in\mathcal{P}_b : \bar\pi_P \ge lo] \,\Vert\, [P_{\text{cur}}]$ \hfill $\triangleright$ compaction
\end{algorithmic}
\end{algorithm}

\paragraph{Bisection instead of sorting.}
Sorting the pages of every row at every layer and step erases most of the saving: in a prototype, a sort-based rule
cut the speedup over dense attention from 3.18$\times$ to 1.47$\times$ (Appendix~\ref{app:design}). \ksel therefore searches for $\tau_b \in [0, \max_P\bar\pi_P]$ directly
(Algorithm~\ref{alg:select}). Each pass tries $K_B$ evenly spaced thresholds, sums the mass above each, and keeps the
sub-interval containing $\tau_b$, shrinking it $K_B$-fold; $24/\log_2 K_B$ passes shrink it $2^{24}$-fold. One thread
block handles one row with all masses in registers, writing only the final page list (sink first, current page last)
and its length $n_b$ to GPU memory.

\subsection{Self-planning attention}
\label{sec:balanced}
\paragraph{The constraint.} A CUDA graph fixes each kernel's thread-block count at capture; at replay only GPU memory
may change. The lengths $n_b$ change every step, so the kernel must handle every length with the same thread blocks,
without the host (Section~\ref{sec:motivation-kernel}).

\paragraph{The idea: each block computes its own share of the work.}
\katt always launches $J$ thread blocks per row. Block $(b,j)$ reads $n_b$ from GPU memory, cuts the row into chunks of
$c_b$ pages, and processes chunk $j$; if the row has no chunk $j$, the block exits at once
(Algorithm~\ref{alg:attend}):
\begin{equation}
J = \max\!\Big(N_J,\ \Big\lceil \frac{N_T}{B H_{kv}} \Big\rceil\Big), \qquad
c_b = \max\!\Big(C,\ \Big\lceil \frac{n_b}{J}\Big\rceil\Big).
\label{eq:items}
\end{equation}
$J$ depends only on the batch size, fixed within each graph; $c_b$ depends on $n_b$, which each block reads itself. For example, with $J{=}4$ and $C{=}2$, every chunk holds two pages, and four rows of 8, 3, 5 and 2
pages use 4, 2, 3 and 1 blocks, and the other six blocks exit immediately (Figure~\ref{fig:overview}, Step~3). Every
busy block processes at most two pages, so the hardware scheduler distributes them evenly over the SMs, with no padding and no plan.
$N_J$ is the minimum number of blocks per row, $N_T$ the minimum for the whole grid (so a few long rows still fill the
GPU), and $C$ the minimum chunk size (so a short row needs one block). Each block computes all $G$ query heads on the
same key/value tile, reading every KV byte once per group. \kcmb combines each row's chunks through their
log-sum-exp; single-chunk rows are written by \katt directly. We use $C{=}24$, $N_J{=}32$ and $N_T{=}4096$ for every model
and shape, and $K_B{=}8$, 4 and 2 for rows of fewer than 256, 1{,}024 and 4{,}096 pages in \ksel.

\begin{algorithm}[t]
  \caption{Self-planning attention: thread block $(b, j)$ of \katt}
  \label{alg:attend}
  \small
\begin{algorithmic}[1]
  \STATE {\bfseries Input:} queries $q_{1..G}$ of row $b$; page list $L_b$ and its length $n_b$ in GPU memory
  \STATE $c \gets \max(C, \lceil n_b/J \rceil)$;\ \ $m \gets \lceil n_b / c\rceil$ \hfill $\triangleright$ chunks of row $b$
  \IF{$j \ge m$}
    \STATE \textbf{return} \hfill $\triangleright$ no chunk for this block
  \ENDIF
  \FOR{$i = jc$ \textbf{to} $\min((j{+}1)c,\, n_b) - 1$}
    \STATE load keys and values of page $L_b[i]$
    \STATE update the attention of all $G$ query heads
  \ENDFOR
  \STATE \textbf{if} $m = 1$ \textbf{then} write the output \textbf{else} write the partial output and its log-sum-exp (LSE)
\end{algorithmic}
\end{algorithm}

\subsection{Overhead analysis}
\label{sec:overhead}
A page summary holds $2D$ values, compared with $2SD$ for the page's keys and values, so it adds $1/S = 1.6\%$ to the
KV-cache size.
Per step, \kest reads all summaries ($1/S$ of the dense bytes) and \katt the kept fraction $\rho$ of the pages, about
$(\rho + 1/S)$ of dense in total; at 128k with $\gam{=}0.98$, $\rho \approx 0.20$, so a step reads about 21\%. \kest and \ksel together take
19--27\,$\mu$s per layer (Table~\ref{tab:nsys}), with no host work.

\subsection{Fixed budgets and other selectors}
\label{sec:generality}
\katt and \kcmb do not depend on how pages are chosen. \ksel also has a \emph{fixed-count mode} that keeps the $K$
largest pages through the same bisection (used in the ablation, Section~\ref{sec:eval-ablation}), and \katt accepts the
page list of any selector through an \texttt{attend()} entry point. On the same kernels we have run a Quest-style upper
bound, a sink-plus-local-window selector in the style of DuoAttention~\citep{xiao2025duoattention}, and blocks of 1 to
128 tokens. Because Quest also stores two vectors per page, running it on the same kernels as \sys isolates the effect of which
pages are kept
(Section~\ref{sec:eval-comp}).

\section{Implementation}
\label{sec:impl}

\paragraph{Integration with vLLM.}
\sys is a Python module of Triton kernels, loaded into vLLM~0.30 through an import hook. It replaces the decode
attention of the FA3 backend and changes neither the model code nor the KV cache layout. The kernels address vLLM's
bfloat16 (bf16) KV cache through its tensor strides as an array [page, KV head, 64 tokens, $2D$], so they work with both memory
layouts that vLLM supports: token-major (NHD, the default) and head-major (HND).

\paragraph{Decode steps and mixed steps.}
In a pure decode step, FA3 is skipped entirely: the four \sys kernels run instead and are captured in vLLM's full CUDA
graph. In a \emph{mixed step}, where some requests are prefilling and others are decoding, \sys handles the decode rows
and FA3 still handles the prefill rows. So that FA3 does not recompute the decode rows, \sys passes FA3 a copy of the
step metadata in which the sequence lengths of the decode rows are zero; FA3 skips them and computes the prefill rows
exactly as dense vLLM does.

\paragraph{Memory for page summaries.}
The summaries ($\mu_P$ and $\sigma_P^2$, bf16) live in a separate buffer, allocated on first use, outside the memory
that vLLM reserves for the KV cache. The KV capacity that vLLM reports is therefore almost unchanged: it differs by only
0.06\% (Llama-3.1-8B: 840{,}320 vs.\ 839{,}808 tokens; Qwen3-4B: 802{,}368 vs.\ 801{,}920). Placing the summaries
inside the KV cache would reduce the capacity by 1.6\%.

\paragraph{Correctness checks.}
At $\gam{=}1$, \sys matches FA3 to within one bf16 unit in the last place (ulp), CUDA-graph replay is bit-identical to eager execution, and the
page lists of \ksel match a float32 reference of the rule on 64/64 test rows; Appendix~\ref{app:correct} gives the
details.

\section{Evaluation}
\label{sec:eval}

\newcommand{\rqitem}[2]{\par\vspace{1pt}\noindent\textbf{#1}\ #2}
Our experiments are driven by the following research questions (RQs):
\rqitem{RQ1. How does \sys compare with prior methods?}{We compare accuracy and latency at 128k with dense attention,
Quest, Twilight and FFD.}
\rqitem{RQ2. Does an adaptive budget beat a fixed one?}{We keep the estimator and kernels of \sys and replace the
coverage rule by fixed, proportional and profiled budgets at equal reads.}
\rqitem{RQ3. Does the kernel keep its speed when page lists differ in length?}{We vary the spread of row lengths at a
fixed mean and compare \katt with FA3, padding, bucketing and FlashInfer.}
\rqitem{RQ4. How much faster is a whole decode step?}{We measure TPOT in vLLM from 8k to 130k tokens and break one
step into kernels.}
\rqitem{RQ5. Does \sys raise throughput in online serving?}{We serve LongBench prompts in a closed loop at increasing
concurrency.}
\par\smallskip

\subsection{Datasets}
\label{sec:eval-data}
{\setlength\topsep{2pt}
\begin{itemize}\setlength\itemsep{1pt}\setlength\parsep{0pt}\setlength\parskip{0pt}
\item \textbf{RULER-13}~\citep{hsieh2024ruler} at 128k (13 tasks $\times$ 20 documents): long-context retrieval (RQ1).
\item \textbf{AIME 2024+2025} (60 problems $\times$ 8 samples, $t{=}0.6$, top-$p$ 0.95, top-$k$ 20, at most 32{,}768
generated tokens): long reasoning (RQ2).
\item \textbf{LongBench}~\citep{bai2024longbench} (21 tasks, 4{,}750 samples): shorter prompts. Its documents also
supply the timing prompts: batches of 8k$\times$32 to 130k$\times$4 tokens (RQ4) and two serving sets of 4k--126k
tokens (RQ5).
\end{itemize}}

\subsection{Evaluation metrics}
\label{sec:eval-metrics}
{\setlength\topsep{2pt}
\begin{itemize}\setlength\itemsep{1pt}\setlength\parsep{0pt}\setlength\parskip{0pt}
\item \textbf{Accuracy}: the benchmark score; differences from dense are tested with a paired bootstrap (95\%).
\item \textbf{Token KL}: KL divergence of a method's token distribution from dense on teacher-forced dense answers.
\item \textbf{Page reads}: fraction of KV pages read relative to dense, counted on the GPU. Quest and Twilight count
reads in tokens and FFD in 128-token blocks, so for them we compare accuracy and time only.
\item \textbf{Latency}: \emph{TPOT}, the time per output token (in serving, the median over requests); \emph{Attn}, the selection plus attention time
of one decode step; \emph{Host}, the CPU planning time per step.
\item \textbf{Throughput}: output tokens per second during serving.
\end{itemize}}

\subsection{Baselines}
\label{sec:eval-base}
Dense FA3 (vLLM's default) and dense FlashInfer; Quest~\citep{tang2024quest} with a 36{,}864-token budget (Quest-36k);
Twilight~\citep{lin2025twilight} with a 36{,}864-token Quest pre-filter and top-$p$ 0.85 (Twilight-36k);
FFD~\citep{liu2026ffd} with its default $\delta{=}7$, each with its authors' code on the same H200 (Quest through
Twilight's implementation of its rule). \sys uses
$\gam{=}0.98$ unless noted, with every layer and KV head sparse.

\subsection{Models and system}
\label{sec:eval-setup}
We use Llama-3.1-8B-Instruct~\citep{grattafiori2024llama3} (32 layers, 32 query heads, 8 KV heads, $D{=}128$) and
Qwen3-4B-Instruct-2507~\citep{yang2025qwen3} (36 layers), and Qwen3-4B-Thinking-2507 for long reasoning. All runs use
one NVIDIA H200 (141\,GB) and vLLM~0.30 with the FA3 attention backend, bf16 weights and KV cache, 64-token pages and
full CUDA graphs; all timings are taken on an otherwise idle GPU. We write 128k for RULER's 131{,}072-token length
(131k in Figure~\ref{fig:cv} and Table~\ref{tab:cv}) and 130k for the 130{,}048-token timing prompts.

\subsection{Experiment results}
\label{sec:eval-results}

\subsubsection{RQ1: How does \sys compare with prior methods?}
\label{sec:eval-acc}

\begin{table}[t]
\caption{\textbf{Accuracy and latency at 128k.} RULER-13 (13 tasks $\times$ 20 documents); $^\dagger$: significantly
below dense (paired bootstrap by document, 95\%). KL: mean token KL to dense on 65 teacher-forced RULER answers (0.0003
with every page kept). Attn: selection\,+\,attention of one decode step at 4$\times$130k on the GPU; Host: FlashInfer's
\texttt{plan()} call per step, with its index arrays already built. TPOT: measured end to end in vLLM with CUDA graphs
on RULER prompts (batch 4); --: the method does not run in vLLM; dense FlashInfer: 18.8 / 20.0\,ms. $\times$: dense
FA3 TPOT / TPOT. Page reads of \sys (and of the \textsuperscript{c} rows, same lists) at $\gam{=}0.98$ / 0.99:
19.6 / 27.4\% on Llama-3.1-8B, 24.2 / 32.5\% on Qwen3-4B. Bold: best sparse value.}
\label{tab:main}
\centering\small
\setlength{\tabcolsep}{2.4pt}
\begin{tabular}{@{}lrrrrrr@{}}
\toprule
& \multicolumn{2}{c}{accuracy} & \multicolumn{4}{c}{latency (ms)} \\
\cmidrule(lr){2-3}\cmidrule(l){4-7}
Method & RULER$\uparrow$ & KL$\downarrow$ & Attn & Host & TPOT & $\times$ \\
\midrule
\multicolumn{7}{@{}l}{\emph{Llama-3.1-8B}} \\
\emph{dense FA3}               & \emph{71.51} & --    & \emph{15.59} & --   & \emph{19.0} & \emph{1.00} \\
Quest-36k\textsuperscript{a}   & 69.63        & 0.022 & 21.98 & 2.11 & --   & --   \\
Twilight-36k                   & 67.94$^\dagger$ & 0.053 & 14.73 & 6.96 & --   & --   \\
FFD $\delta7$\textsuperscript{b} & 64.85$^\dagger$ & 0.071 & 13.03 & --   & 18.8 & 1.01 \\
FlashInfer\textsuperscript{c}  & 69.60        & 0.022 & \textbf{4.34} & 2.16 & 12.6 & 1.51 \\
FA3\textsuperscript{c}         & 69.60        & 0.022 & 4.86  & --   & 7.6  & 2.50 \\
\rowcolor{oursrow} \sys $\gam{=}0.98$ & 69.60 & 0.022 & 4.73 & -- & \textbf{7.1} & \textbf{2.69} \\
\rowcolor{oursrow} \sys $\gam{=}0.99$ & \textbf{71.83} & \textbf{0.015} & 6.43 & --   & 7.7  & 2.46 \\
\midrule
\multicolumn{7}{@{}l}{\emph{Qwen3-4B}} \\
\emph{dense FA3}               & \emph{83.62} & --    & \emph{17.53} & --   & \emph{20.2} & \emph{1.00} \\
Quest-36k\textsuperscript{a}   & 82.69        & 0.020 & 24.71 & 2.29 & --   & --   \\
Twilight-36k                   & 82.26        & 0.056 & 16.26 & 7.40 & --   & --   \\
FFD $\delta7$\textsuperscript{b} & \textbf{83.78} & 0.015 & 16.28 & --   & --   & --   \\
FlashInfer\textsuperscript{c}  & 83.54        & 0.018 & \textbf{5.86} & 2.25 & 13.2 & 1.53 \\
FA3\textsuperscript{c}         & 83.54        & 0.018 & 6.55 & --   & 8.6  & 2.36 \\
\rowcolor{oursrow} \sys $\gam{=}0.98$ & 83.54 & 0.018 & 6.63 & -- & \textbf{7.6} & \textbf{2.68} \\
\rowcolor{oursrow} \sys $\gam{=}0.99$ & 83.70 & \textbf{0.014} & 8.06  & --   & 9.0  & 2.25 \\
\bottomrule
\end{tabular}
\\[2pt]\parbox{\columnwidth}{\footnotesize \textsuperscript{a}Quest's rule as implemented in Twilight's code, not the
original Quest kernels: Quest bound and top-$k$ per query head, then FlashInfer over Quest's own 16-token pages, one 36{,}864-token list per query head; Host: FlashInfer's
plan for these lists. \textsuperscript{b}Fused scan and attention at FFD's RULER kept fraction (0.125 / 0.212 of the
128-token blocks), 32 / 36 layers; TPOT measured in FFD's own runtime (Hugging Face with full CUDA graphs), which
supports Llama only. \textsuperscript{c}The $\gam{=}0.98$ page lists of \sys, attended by FlashInfer's
tensor-core paged kernel or by FA3 with each (request, KV head) as a varlen sequence: same pages, hence same accuracy.
Attn: FA3's split metadata precomputed and not timed; Host: FlashInfer's plan per step. In the TPOT run, FA3 builds its
split metadata in every layer, and FlashInfer, which cannot plan inside a CUDA graph, reuses one plan made for rows of
full width.}
\end{table}

To answer this RQ, we report RULER and token KL at 128k next to Attn, Host and TPOT in Table~\ref{tab:main}. We time the Attn
of every method on the same GPU and batch shape, each at the budget of its RULER row. TPOT is measured end to end with
CUDA graphs: in vLLM for \sys, FA3\textsuperscript{c} and FlashInfer\textsuperscript{c}, and in FFD's own runtime for
FFD; Quest and Twilight do not run in vLLM, so we compare them by Attn and Host (protocol in
Appendix~\ref{app:comp}).

\paragraph{On Llama-3.1-8B, only \sys is both as accurate as dense and faster than dense.}
Every prior method gives up one of the two there. Twilight and FFD lose significant accuracy (67.94 and 64.85
vs.\ 71.51). Quest keeps its accuracy, but its attention is slower than dense; with their host plans, both Quest and
Twilight spend more time on attention per step than dense FA3 (24.1 and 21.7 vs.\ 15.6\,ms on Llama-3.1-8B). FFD's
attention is 1.20$\times$ faster than dense, but its measured TPOT equals dense (18.8 vs.\ 19.0\,ms). On Qwen3-4B,
FFD keeps dense accuracy and its attention is 1.08$\times$ faster than dense; its runtime does not support this model,
so its TPOT is not measured. \sys at $\gam{=}0.99$ matches dense on both models (71.83 vs.\ 71.51, 83.70 vs.\ 83.62)
and reduces TPOT by \textbf{2.2--2.5$\times$}; at $\gam{=}0.98$, it is 2.7$\times$ faster than dense and still within
the noise of dense.

\paragraph{Against each method.}
On Llama-3.1-8B, $\gam{=}0.99$ scores 3.9 points above Twilight-36k and 7.0 above FFD, both significant. FFD's TPOT
is 2.4$\times$ that of $\gam{=}0.99$ (18.8 vs.\ 7.7\,ms), and Twilight's attention step with the host plan takes
3.4$\times$ as long (21.7 vs.\ 6.4\,ms). $\gam{=}0.98$ matches
Quest-36k with only 19.6\% of the pages, about 28\% fewer than Quest's budget; at similar reads, $\gam{=}0.99$ scores 2.2
points above Quest (not significant). On Qwen3-4B, every method is within the noise of dense. Token KL agrees: Twilight (on both
models) and FFD (on Llama-3.1-8B) are farther from dense than \sys, while Quest is on par.

\label{sec:eval-comp}
\paragraph{Each prior method loses for one main reason.}
\emph{Quest's rule} selects per query head, so FlashInfer's paged kernel attends over 32 lists instead of 8 (8.30 vs.\
4.89\,ms with the CUDA-core kernel, plus a 2.16\,ms host plan); it also needs 27.4\% of the pages where $\gam{=}0.98$
needs 19.6\% at equal accuracy, and replacing it by the coverage rule alone makes attention 1.36$\times$ faster. \emph{FFD} scans a 2-bit copy of every key at every step,
so keeping 12.5--21.2\% of the blocks makes its attention only 1.08--1.20$\times$ faster than dense, and on
Llama-3.1-8B its measured TPOT equals dense; the attention path of
\sys is \textbf{1.70--4.19$\times$} faster than FA3 (Table~\ref{tab:nsys}). \emph{Twilight} treats GQA as 32
independent heads, so its selection chain alone takes 12.71\,ms per step, plus a 6.96\,ms host plan outside any CUDA
graph.

\paragraph{At 128k, the speedup comes mainly from reading fewer pages.}
The FlashInfer\textsuperscript{c} and FA3\textsuperscript{c} rows run existing kernels on exactly the page lists of
\sys $\gam{=}0.98$, which separates the effect of the rule from that of the kernel. At 128k there are only 32 rows,
few enough for existing kernels to balance on their own: in isolation, FA3 is within 3\% of \katt, and FlashInfer's
tensor-core kernel is 11--15\% faster. End to end, FA3\textsuperscript{c} is 7--14\% slower than \sys (7.6 vs.\
7.1\,ms on Llama-3.1-8B, 8.6 vs.\ 7.6\,ms on Qwen3-4B), a gap that includes building FA3's split metadata in every
layer. FlashInfer cannot plan inside a CUDA graph: with a plan per layer and step, TPOT on Llama-3.1-8B is 24.7\,ms,
and inside the graph it must reuse one plan made for rows of full width, which brings TPOT to 12.6\,ms (13.2\,ms on
Qwen3-4B). At this shape the speedup of \sys over dense therefore comes mainly from the rule; the kernel pays off when
there are many rows of uneven length (Section~\ref{sec:eval-kernel}) and when the lists change at every step.

\paragraph{The output distribution stays close to dense.}
We teacher-force the dense answers on 65 RULER documents at 128k (5 per task) and compare every token's distribution
with every layer sparse (1{,}664 tokens on Llama-3.1-8B, 2{,}437 on Qwen3-4B). Mean token KL is 0.022 at
$\gam{=}0.98$ and 0.015 at $\gam{=}0.99$ on Llama-3.1-8B (Qwen3-4B: 0.018 and 0.014), compared with 0.0003 (numerical
error) when every page is kept. On Llama-3.1-8B, variable tracking accounts for only 43 tokens but a third of the KL,
and for most of the RULER loss at $\gam{=}0.98$. On LongBench (all 4{,}750 samples), whose prompts are shorter, Llama-3.1-8B scores 49.78 dense and 49.90 at
$\gam{=}0.98$ (33.7\% reads); Qwen3-4B scores 47.59 and 47.31 (38.6\% reads).

\subsubsection{RQ2: Does an adaptive budget beat a fixed one?}
\label{sec:eval-ablation}

To answer this RQ, we keep the
estimator and kernels of \sys, change only the rule that sets each row's page count, and compare three kinds of fixed budget:
\emph{fixed $K$}: every row keeps the $K$ pages with the largest estimated mass, and rows shorter than $K$ are read in
full (the fixed-count mode of \ksel, Section~\ref{sec:generality});
\emph{proportional top-$k$}: every row keeps the same fraction of its pages, chosen so that the total reads equal those
of $\gam{=}0.98$;
\emph{profiled $K$}: we run $\gam{=}0.98$ on AIME 2022--2023 (60 problems, two solutions each, disjoint from the test
problems), record how many pages each layer, or each KV head of each layer, keeps, and fix one count per layer (36
counts, 50--204 pages) or per (layer, KV head) (288 counts, 8--318 pages), scaled so that the total reads match those of
$\gam{=}0.98$.

\paragraph{Why AIME rather than RULER.}
On RULER the answer is a few tokens long and the needed information sits in a few pages; once the budget is large
enough, every rule keeps them. Fixed $K$ set to the reads of $\gam{=}0.98$ and $0.99$ is within 0.35 points of the
coverage rule on both models, far below the benchmark noise (about 1.3--2.7 points), and these $K$ are known only after
the coverage rule has found the right reads. On AIME the model generates up to 32{,}768 reasoning tokens and every
step reads that reasoning back; small per-step errors accumulate over thousands of steps, so AIME exposes
under-budgeted steps that RULER misses.

\begin{table}[t]
\caption{\textbf{Adaptive vs.\ fixed budgets on AIME} (Qwen3-4B-Thinking, 480 solutions). All sparse arms use the same
estimator and kernels and differ only in the rule that sets the page count. Truncated: solutions that hit the
32{,}768-token limit. Closed acc.: accuracy on the solutions that finish before the limit. Accuracy also counts the
0--2 truncated solutions per arm whose answer, extracted from the truncated text, is correct. $^\dagger$: significantly
below \sys $\gam{=}0.98$ (paired bootstrap by problem, 95\%).}
\label{tab:ablation}
\centering\small
\setlength{\tabcolsep}{3pt}
\begin{tabular}{@{}lrrrr@{}}
\toprule
            & Pages     & Accuracy & Trun-  & Closed \\
Budget      & read (\%) & (\%)     & cated  & acc. (\%) \\
\midrule
\emph{dense} & \emph{100} & \emph{75.2} & \emph{89} & \emph{92.1} \\
\addlinespace[3pt]
\multicolumn{5}{@{}l}{\textit{Fixed budget}} \\
\hspace{0.6em}proportional top-$k$ & 39.5 & 41.0$^\dagger$ & 231 & 78.3 \\
\hspace{0.6em}fixed $K{=}75$     & 33.1 & 69.8$^\dagger$ & 131 & 95.7 \\
\hspace{0.6em}fixed $K{=}90$     & 39.8 & 72.7           & 108 & 93.8 \\
\hspace{0.6em}profiled $K$, per layer   & 38.7 & 72.5 & 112 & 94.3 \\
\hspace{0.6em}profiled $K$, per KV head & 39.7 & 74.6 &  93 & 92.0 \\
\addlinespace[3pt]
\multicolumn{5}{@{}l}{\textit{Adaptive (coverage rule)}} \\
\rowcolor{oursrow}\hspace{0.6em}\sys $\gam{=}0.98$ & 39.4 & 75.2          & 94 & 93.5 \\
\rowcolor{oursrow}\hspace{0.6em}\sys $\gam{=}0.99$ & 49.9 & \textbf{75.4} & 91 & 93.1 \\
\bottomrule
\end{tabular}
\end{table}

\paragraph{At equal reads, proportional top-$k$ loses almost half of the accuracy.}
Proportional top-$k$ reads as much as \sys (39.5\% vs.\ 39.4\%) but reaches only 41.0\% against 75.2\%
(Table~\ref{tab:ablation}). Almost half of its solutions (231/480) hit the length limit, and those that finish are correct
less often (78.3\% vs.\ 93.5\%). With one fraction for every row, heads with concentrated attention over-read while
heads with diffuse attention lack pages at many steps.

\paragraph{Fixed $K$ hurts less but depends on choosing the right $K$.}
Under fixed $K$ only rows longer than $K$ run short of budget. At $K{=}75$, the mean page count of $\gam{=}0.98$,
accuracy falls to 69.8\%, entirely through truncation (131 vs.\ 94 solutions hit the limit), while finished solutions
are as accurate as with \sys (95.7\% vs.\ 93.5\%): a short budget does not make the model answer wrongly; it makes
the model reason longer until it reaches the length limit. $K{=}90$, at the reads of \sys, reaches 72.7\%, within noise, yet is
still truncated more often (108 vs.\ 94). A fixed $K$ works only when tuned per model and task; the coverage rule uses
the same two settings of $\gam$ for every model and task and adapts the page count to every row and step.

\paragraph{A profiled budget comes close only per KV head and after a calibration run.}
Profiling makes the fixed budget follow the variation across layers and heads, though not across steps. One count
per layer reaches 72.5\% and is truncated more often (112 vs.\ 94). One count per (layer, KV head) reaches 74.6\%,
within noise of the coverage rule (75.2\%), with as many truncated solutions. The coverage rule stays at or above
every static budget at the same reads, and it needs no profiling run: the 288 counts are fitted to this model and this
task, whereas this paper uses the same two settings of $\gam$ for every model and task.

\paragraph{\sys matches dense on long reasoning.}
$\gam{=}0.98$ and $0.99$ reach 75.2\% and 75.4\%, equal to dense (75.2\%), reading 39.4\% and 49.9\% of the pages.
About a fifth of the solutions hit the 32k limit under both dense and \sys, so the limit, not the attention, accounts
for most of the remaining error. The 480 solutions also finish 1.86$\times$ sooner (2{,}748 vs.\ 5{,}110\,s). When
AIME prompts (20.4k output tokens on average) are served in vLLM, $\gam{=}0.98$ raises output throughput by
19\%, 67\% and 88\% at concurrency 8, 32 and 64 (one run per point).

\paragraph{The coverage rule stays closer to dense and does not slow decoding.}
On 60 LongBench requests with Llama-3.1-8B at equal reads, the coverage rule's token KL to dense is about 0.37 times
that of proportional top-$k$, and lower on every request; on all 4{,}750 LongBench samples the scores tie (49.90 and 49.88 at 33.7\% reads). Against fixed $K$
at equal reads, TPOT differs by $-5.5\%$ to $+6.4\%$ with no consistent sign, so a per-row budget has no consistent time cost beyond
run-to-run variation (Appendix~\ref{app:extra}).

\subsubsection{RQ3: Does the kernel keep its speed when page lists differ in length?}
\label{sec:eval-kernel}

\begin{figure*}[t]
  \centering
  \includegraphics[width=\textwidth]{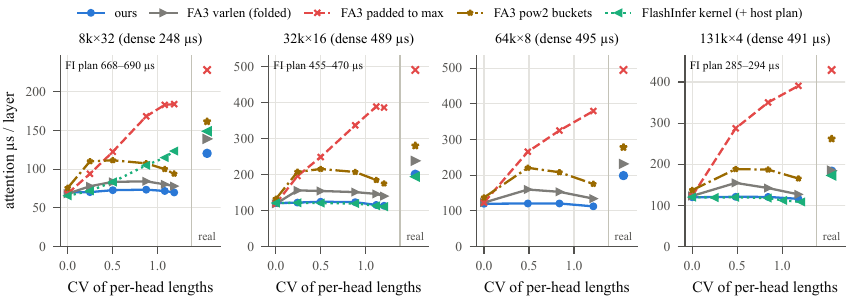}
  \vspace{-6mm}
  \caption{\textbf{Attention time per layer as per-head lengths spread} (Llama-3.1-8B geometry, 22\% of the pages kept
  on average; the x axis is the coefficient of variation (CV) of the per-(request, KV head) lengths). ``real'': per-(request,
  layer, KV head) page counts from decode steps of \sys on LongBench prompts of about 8k tokens, applied as fractions
  of each context (0.33--0.41 of the pages; short contexts need a larger fraction than RULER at 128k). FlashInfer additionally needs the host plan given in each panel,
  per layer and step. The 64k panel has no FlashInfer arm.}
  \label{fig:cv}
\end{figure*}

To answer this RQ, we keep 22\% of the pages on average and vary only the coefficient of variation (CV, standard
deviation over mean) of the row lengths (Figure~\ref{fig:cv}); ``real'' uses page counts from decode steps of \sys at about 8k tokens, applied as fractions of each context. All arms
run on the same lists:
\emph{FA3-varlen} treats each row as its own sequence, \emph{padding} extends every row to the longest one,
\emph{bucket} groups rows by powers of two, and \emph{FlashInfer} needs a host \texttt{plan()} call before each step.

\paragraph{The time of \sys is nearly flat in CV.}
At 8k$\times$32 and 32k$\times$16, \sys stays within 69.9--73.5 and 113.2--124.5\,$\mu$s per layer as CV grows from 0 to
1.2, while FA3-varlen slows by 23\% and 32\%, padding by 2.7$\times$ and 3.3$\times$, and buckets by 1.5$\times$ and
1.6$\times$.
On real page counts at 8k and 32k, FA3-varlen is 1.16$\times$ and 1.19$\times$ slower than \sys, padding 1.90$\times$
and 2.45$\times$, and buckets 1.34$\times$ and 1.40$\times$.
At 131k$\times$4 there are only 32 rows: on real lists FA3-varlen balances them on its own and is as fast as \sys
(184.5 vs.\ 183.0\,$\mu$s), but at CV 1 it is 18\% slower (142.0 vs.\ 120.4\,$\mu$s, Table~\ref{tab:cv}), while \sys stays flat in CV thanks to the $N_T$ floor of Eq.~(\ref{eq:items}).

\paragraph{FlashInfer's kernel is about as fast; its host plan is not.}
FlashInfer's kernel is 1.24$\times$ slower than \sys at 8k, 3\% faster at 32k and 6\% faster at 131k. Its host plan
must run before every layer of every step, outside the CUDA graph. Measured end to end here (copying the page counts
to the host, building the index arrays and calling \texttt{plan()}), it takes 285--690\,$\mu$s per layer, 9--22\,ms over
32 layers, longer than a whole decode step of \sys (7.3--10.8\,ms, Table~\ref{tab:tpot}). The \texttt{plan()} call
alone, with the index arrays already built, takes 2.2\,ms per step at 131k$\times$4 (Table~\ref{tab:main}).
\paragraph{Why not one shared list for all KV heads.}
A list shared by all KV heads of a (request, layer) would let existing paged kernels run unchanged, but it must hold
the union of every head's pages: 78--87\% of the pages on real data, against 33--41\% with per-head lists
(Appendix~\ref{app:cv}).

\subsubsection{RQ4: How much faster is a whole decode step?}
\label{sec:eval-tpot}

\begin{table}[t]
\caption{\textbf{TPOT (ms)} of \sys ($\gam{=}0.98$) and dense FA3 on prompts cut from LongBench documents (batch
$\times$ context as listed), mean of two interleaved pairs (whose speedups differ
by at most 0.13), and page reads of \sys. Bold: lower TPOT.}
\label{tab:tpot}
\centering\small
\setlength{\tabcolsep}{2.2pt}
\begin{tabular}{@{}lrrrrrrrr@{}}
\toprule
& \multicolumn{4}{c}{Llama-3.1-8B} & \multicolumn{4}{c}{Qwen3-4B} \\
\cmidrule(lr){2-5}\cmidrule(l){6-9}
Shape & dense & \sys & $\times$ & reads & dense & \sys & $\times$ & reads \\
\midrule
8k$\times$32   & 12.66 & \textbf{9.18}  & 1.38 & 0.42 & 12.42 & \textbf{8.98}  & 1.38 & 0.49 \\
16k$\times$32  & 19.24 & \textbf{10.83} & 1.78 & 0.37 & 20.59 & \textbf{11.68} & 1.76 & 0.43 \\
32k$\times$16  & 16.46 & \textbf{8.71}  & 1.89 & 0.31 & 17.00 & \textbf{8.92}  & 1.91 & 0.39 \\
64k$\times$8   & 19.56 & \textbf{7.45}  & 2.62 & 0.16 & 20.43 & \textbf{8.39}  & 2.44 & 0.25 \\
130k$\times$4  & 19.32 & \textbf{7.32}  & \textbf{2.64} & 0.15 & 20.26 & \textbf{7.53}  & \textbf{2.69} & 0.21 \\
\bottomrule
\end{tabular}
\end{table}

\begin{table}[t]
\caption{\textbf{One decode step of Llama-3.1-8B by kernel} (ms, Nsight Systems median of 128 steps). att, est, sel,
mrg: \katt, \kest, \ksel, \kcmb; GEMM: matrix multiplications. Tracing at CUDA-graph node level lengthens
the step by up to 1\,ms relative to Table~\ref{tab:tpot}. Bold: shorter step.}
\label{tab:nsys}
\centering\small
\setlength{\tabcolsep}{3pt}
\begin{tabular}{@{}lrrrrrrrr@{}}
\toprule
& \multicolumn{2}{c}{dense} & \multicolumn{6}{c}{\sys $\gam{=}0.98$} \\
\cmidrule(lr){2-3}\cmidrule(l){4-9}
Shape & step & FA3 & step & att & est & sel & mrg & GEMM \\
\midrule
8k$\times$32   & 12.72 & 8.16  & \textbf{9.38} & 4.02 & 0.38 & 0.23 & 0.18 & 3.96 \\
32k$\times$16  & 17.00 & 12.48 & \textbf{9.45} & 4.20 & 0.40 & 0.25 & 0.13 & 3.86 \\
64k$\times$8   & 20.34 & 15.89 & \textbf{8.44} & 3.04 & 0.44 & 0.41 & 0.12 & 3.83 \\
130k$\times$4  & 20.16 & 15.76 & \textbf{8.18} & 2.78 & 0.43 & 0.42 & 0.13 & 3.83 \\
\bottomrule
\end{tabular}
\end{table}

To answer this RQ, we prefill synthetic batches
with 8k--130k-token contexts, decode 256 tokens and measure TPOT in vLLM (Table~\ref{tab:tpot};
Figure~\ref{fig:tpot} in the appendix). Table~\ref{tab:nsys} breaks one step into kernels with Nsight Systems, traced at
CUDA-graph node level.

\paragraph{The longer the context, the larger the speedup over dense.}
TPOT falls by 1.38$\times$ at 8k$\times$32 and by 2.64$\times$ (Llama-3.1-8B) and 2.69$\times$ (Qwen3-4B) at
130k$\times$4: with longer contexts attention takes a larger share of the step and the kept fraction falls (for
Llama-3.1-8B, from 42\% at 8k to 15\% at 130k). With $\gam{=}0.99$, which matches dense on RULER, TPOT falls by
1.25--2.27$\times$ on Llama-3.1-8B and 1.23--2.27$\times$ on Qwen3-4B over the same shapes
(Table~\ref{tab:tpot99} in the appendix).

\paragraph{Selection takes about a tenth of the step or less.}
The GEMMs take 3.8--4.0\,ms under both, so the whole difference lies in attention. FA3 grows from 8.16 to
15.76\,ms with the context, while the four kernels of \sys stay at 3.8--5.0\,ms: the attention path is
1.70$\times$ faster than FA3 at 8k$\times$32 and 4.19$\times$ at 130k$\times$4. \kest and \ksel together take only
0.61--0.85\,ms per step. In mixed steps, prefill rows still run through FA3 unchanged; with continuous batching on
LongBench, output throughput rises by 18\% (Appendix~\ref{app:extra}).

\subsubsection{RQ5: Does \sys raise throughput in online serving?}
\label{sec:eval-serve}

\begin{table}[t]
\caption{\textbf{Online serving} (\texttt{vllm bench serve}, closed loop, 1{,}024 output tokens). Set A: 256 LongBench
prompts of 4k--64k tokens; set B: 64 prompts of 64k--126k tokens. $c$: concurrency. FI: dense FlashInfer. Bold: best
per row.}
\label{tab:serve}
\centering\small
\setlength{\tabcolsep}{2.8pt}
\begin{tabular}{@{}lrrrrrrr@{}}
\toprule
& & \multicolumn{3}{c}{median TPOT (ms) $\downarrow$} & \multicolumn{3}{c}{output tok/s $\uparrow$} \\
\cmidrule(lr){3-5}\cmidrule(l){6-8}
Model & set, $c$ & FA3 & FI & \sys & FA3 & FI & \sys \\
\midrule
\multirow{4}{*}{Llama-8B}
 & A, 1  & \textbf{5.63} & 5.86 & 6.36 & \textbf{152} & 147 & 137 \\
 & A, 16 & 28.6 & 34.2 & \textbf{17.2} & 511 & 446 & \textbf{743} \\
 & B, 1  & 8.16 & 8.18 & \textbf{6.71} & 64.9 & 63.5 & \textbf{71.2} \\
 & B, 8  & 65.5 & 63.6 & \textbf{31.6} & 99.9 & 101.6 & \textbf{142.0} \\
\midrule
\multirow{4}{*}{Qwen3-4B}
 & A, 1  & \textbf{4.37} & 4.62 & 5.10 & \textbf{194} & 185 & 171 \\
 & A, 16 & 25.4 & 32.1 & \textbf{13.9} & 563 & 471 & \textbf{851} \\
 & B, 1  & 7.22 & 7.25 & \textbf{5.57} & 71.5 & 69.1 & \textbf{79.9} \\
 & B, 8  & 65.8 & 63.6 & \textbf{33.9} & 101.1 & 103.0 & \textbf{150.2} \\
\bottomrule
\end{tabular}
\end{table}

To answer this RQ, we run \texttt{vllm bench
serve} in a closed loop with LongBench prompts, prefix caching off and 1{,}024 output tokens per request
(Table~\ref{tab:serve}; Figure~\ref{fig:serve} in the appendix covers every concurrency). Every request completes.

\paragraph{Under high load, throughput rises by about half.}
At concurrency 16 on set A, \sys lowers median TPOT by 40--45\% and raises output throughput by 45\% on Llama-3.1-8B and
51\% on Qwen3-4B. With the long prompts of set B at concurrency 8, \sys halves TPOT on both models, and the mean
end-to-end latency of Llama-3.1-8B falls from 81.8 to 57.6\,s. Dense FlashInfer is never more than 4\% faster than dense
FA3, and $\gam{=}0.99$ keeps most of the gain (Figure~\ref{fig:serve}).

\paragraph{With a single short request, \sys is slower.}
With one request, the GEMMs dominate the step and the four kernels of \sys add an almost constant cost. From 2k to 64k
tokens, TPOT of \sys stays at 6.32--6.50\,ms on Llama-3.1-8B (4.99--5.40\,ms on Qwen3-4B), while dense FA3 grows
from 4.98 to 6.80\,ms (3.63 to 5.76\,ms): \sys is slower up to 32k tokens and 1.05$\times$ (1.07$\times$) faster at
64k. Serving shows the same split: at concurrency 1, \sys is slower on set A and faster on set B. Building the page
summaries during prefill adds 0.2\% to the time to first token (TTFT) of a single 128k prompt (10.60 vs.\ 10.58\,s on Llama-3.1-8B, 10.21
vs.\ 10.19\,s on Qwen3-4B). In a closed loop TTFT does not improve: faster decoding admits new prompts sooner, so more
prefill work queues behind the running batch.

\section{Discussion}
\label{sec:discussion}

\label{sec:qual}
This section examines in more detail how \sys spends its budget in a real run.

\subsection{Where does the budget go?}
In the AIME run of \sys ($\gam{=}0.98$, Qwen3-4B-Thinking), the mean fraction of pages kept per (layer, KV head)
ranges from 3.7\% to 92\%. Layers 0--3 keep 56--74\% of their pages, the sparsest layers about a quarter, and within
one layer the KV head that reads the most keeps a median of 4.0$\times$ as many pages as the one that reads the least
(up to 16.6$\times$). The estimator thus reproduces, in a real run, the variation that Figure~\ref{fig:motivation}a
shows for the oracle sets.

\subsection{Why do fixed budgets fail on long reasoning?}
We group the 60 AIME problems by the median length of their eight dense solutions: 23 are shorter than 16k tokens, 22
longer than 24k, and 15 in between. On the 23 short problems, every arm of Table~\ref{tab:ablation} except proportional top-$k$ solves every sample. On
the 22 long problems, \sys reaches 42.0\% (dense 40.3\%), while every static budget scores
lower: fixed $K$ 31.8--36.4\%, profiled $K$ 34.7--38.6\% and proportional top-$k$ 17.6\%. A few under-budgeted steps
lengthen the reasoning and push it past the length limit. Problem II-5 of AIME 2024 shows this most clearly: dense
solves 6 of 8 samples with a median of 31.5k tokens and \sys 7 of 8, while proportional top-$k$ truncates all 8 samples
and both profiled budgets truncate 6.

\section{Threats to Validity}
\label{sec:threats}

\paragraph{Internal validity.}
Prior methods run with their authors' code and kernels on the same GPU, with the settings of Section~\ref{sec:eval-base}; Quest's
rule runs as implemented in Twilight's code. Quest and Twilight do not run inside vLLM, so we compare them by Attn and
Host only. FFD's TPOT is measured in its own runtime, whose cost outside attention may differ from vLLM's, and with
prompts cut to equal length because its cache holds one length per batch. FlashInfer\textsuperscript{c} runs inside
the CUDA graph with one plan made for rows of full width, our wiring, since planning at every step is not possible
there. The FA3\textsuperscript{c} and FlashInfer\textsuperscript{c} rows use vLLM's head-major (HND) KV-cache layout,
under which \sys is 0.3\,ms slower than in the default token-major (NHD) layout; Table~\ref{tab:main} reports \sys under the same layout. With every page kept,
\sys reproduces dense FA3 (token KL 0.0003; Appendix~\ref{app:correct}).

\paragraph{Construct validity.}
Page reads are counted on the GPU for every (request, layer, KV head); Quest and Twilight select tokens and FFD
128-token blocks, so against them we compare accuracy and time rather than reads. Timings are medians of repeated runs
on an otherwise idle GPU with dense and sparse arms interleaved, and accuracy differences are tested with a paired
bootstrap by document or by problem.

\paragraph{External validity.}
We evaluate 4B and 8B models from two families with a bf16 KV cache on one H200, under vLLM~0.30. On larger models the
GEMMs take a larger share of the step, so the speedup will be smaller. Quantized KV caches, tensor parallelism and
latent attention are not covered, serving is measured in a closed loop only, and \sys saves bandwidth, not capacity:
the full KV cache stays resident.

\section{Related Work}
\label{sec:related}

\paragraph{Fixed budgets.}
Quest~\citep{tang2024quest} bounds the logits of each page with a per-channel key box and keeps a fixed token budget.
CompKV~\citep{huang2026compkv} estimates each block's mass from the mean and variance of its keys, the same
approximation as \mvar, but keeps a fixed number $K$ of blocks and compensates for the dropped ones with the block mean.
InfiniGen~\citep{lee2024infinigen}, ShadowKV~\citep{sun2025shadowkv}, MagicPIG~\citep{chen2025magicpig},
RetrievalAttention~\citep{liu2024retrievalattention} and RetroInfer~\citep{chen2025retroinfer} predict or retrieve
important tokens, often with CPU offload; The Sparse Frontier~\citep{nawrot2025sparsefrontier} evaluates these
trade-offs. Production sparse attention is trained: DeepSeek Sparse
Attention~\citep{deepseek2025v32}, MiniMax Sparse Attention~\citep{lai2026msa} and Qwen3.8-Next~\citep{qiu2026qwen38next}
learn an indexer and keep about 2{,}048 tokens per query, so their kernels never see variable lengths. \sys needs no
training and makes the budget an output of the selection.

\paragraph{Adaptive budgets.}
Twilight~\citep{lin2025twilight} applies top-$p$ to Quest's candidates per query head, attends through FlashInfer with a
host \texttt{plan()} per layer and step, and uses a uniform budget in its end-to-end evaluation. Tactic~\citep{zhu2025tactic}
splits requests into FlashInfer sub-requests. SeerAttention-R~\citep{gao2025seerattentionr} thresholds a learned block
gate per KV head but pads to the longest selection, and PSA~\citep{zhou2025psa} grows its budget with a host loop per
layer. Double-P~\citep{ni2026doublep} uses hierarchical thresholds, SpotAttention~\citep{ahmad2026spotattention} applies
top-$p$ to a learned selector, and Elastic Threshold Attention~\citep{haris2026eta} trains its thresholds; a hybrid CPU--GPU
design~\citep{yao2026hybrid} selects per head across PCIe; vAttention~\citep{desai2025vattention} bounds the error
statistically. FFD~\citep{liu2026ffd} is the closest systems design: its fused Triton kernels need no index lists, scan a
2-bit copy of every key (Section~\ref{sec:eval-comp}), and are captured in a CUDA graph by its own runtime. None of
these methods runs inside the CUDA graphs of a serving engine; \sys brings per-KV-head budgets there at low cost.

\paragraph{Per-head KV eviction.}
Ada-KV~\citep{feng2024adakv}, HeadKV~\citep{fu2025headkv}, DuoAttention~\citep{xiao2025duoattention},
LServe~\citep{yang2025lserve}, KV-Compress~\citep{rehg2024kvcompress} and HARD-KV~\citep{yang2026hardkv} set per-head
budgets once after prefill, offline, or every 128 steps (HARD-KV, inside vLLM's CUDA graphs), mostly on padded grids; H2O~\citep{zhang2023h2o},
StreamingLLM~\citep{xiao2024streamingllm} and SnapKV~\citep{li2024snapkv} evict tokens without looking at the current query. They change
what is stored; \sys changes what is read, at every step.

\paragraph{Decode kernels.}
Flash-Decoding~\citep{dao2023flashdecoding,hong2024flashdecodingpp} and Lean Attention~\citep{sanovar2024lean} split
long sequences across thread blocks. FlashAttention~\citep{dao2022flashattention,dao2024flashattention2,shah2024flashattention3}
chooses splits on the GPU from the batch shape, and FlashInfer~\citep{ye2025flashinfer} balances ragged lists with a
host plan. Our kernel derives an equal-cost partition from per-row lengths in device memory. 

\section{Conclusion and Future Work}
\label{sec:conclusion}
\label{sec:final}
Sparse decode pays off most when the budget follows the query and the kernel follows the budget. \sys selects pages
per (request, layer, KV head, step) with a coverage rule and attends over the ragged lists with a self-planning kernel
inside vLLM's CUDA graphs. On an H200 at 128k, with $\gam{=}0.99$ it matches dense RULER accuracy with 27--33\% of the page reads and
reduces TPOT by \textbf{2.2--2.5$\times$}; on synthetic batches of 8k--130k tokens it lowers TPOT by 1.38--2.69$\times$ at
$\gam{=}0.98$, raises serving throughput by \textbf{45--51\%} under high load, and on long reasoning it matches dense with about 40\%
of the reads.

Several directions remain open. On a quantized KV cache~\citep{liu2024kivi,hooper2024kvquant}, every page shrinks while
the rule reads fewer pages, so the two savings multiply. Under latent attention~\citep{deepseek2024v2}, the rule would
select pages of the shared latent cache. Estimated page masses could tell an offloading system such as
ShadowKV~\citep{sun2025shadowkv} which pages to keep on the GPU or prefetch. Applied to a block of queries, the rule
extends to speculative decoding~\citep{sadhukhan2025magicdec}, which verifies several draft tokens at once; applied to
learned scores, it would make trained sparse attention~\citep{yuan2025nsa} adaptive. A device-side plan for
FlashInfer and a port to SGLang~\citep{zheng2024sglang} are engineering work.

\bibliography{refs}
\bibliographystyle{mlsys2025}

\clearpage
\onecolumn
\raggedbottom  %
\appendix

\section{Design Details}
\label{app:design}
This appendix complements Section~\ref{sec:design}.

\paragraph{Estimator choice.}
We compare page-mass estimators by the token KL to dense when every layer decodes sparsely (Llama-3.1-8B, 60
LongBench requests). With the exact (oracle) mass, top-$p$ at 0.95 keeps 23.9\% of the pages at a KL of 0.00348. An
int4 copy of the keys comes close to the oracle (24.1\% kept, KL 0.00343) but reads an extra 15.6\% of the keys.
\mvar at $\gam{=}0.98$ keeps 37.1\% at a KL of 0.00441, and at $\gam{=}0.995$ keeps 61.7\% at 0.00095. Estimators that
use only the key mean fail on heads that concentrate attention on a few scattered tokens: one KV head of layer 2
accounts for almost all of their KL, because with the mean alone the mass of such a page is underestimated by a
factor of $e^{5}$--$e^{7}$, against about $e^{3}$ with the variance term of Eq.~(\ref{eq:mvar}).

\paragraph{Why pages are not sorted.}
In a prototype that preceded the final module, the coverage rule implemented with a PyTorch sort cost 66\,$\mu$s per
layer at a 6.6k context and erased most of the gain: the whole pipeline was 1.47$\times$ faster than dense
attention, against 3.18$\times$ for the attention kernel alone. Fusing selection into one kernel (\ksel,
Section~\ref{sec:design}) raised this to 2.56$\times$.

\paragraph{Kernel configuration.}
\kest runs several thread blocks per row (at least four per SM over the batch), each covering 64 pages, and writes
the summary of the current page as soon as a new token fills it, so a decode step needs no separate summary kernel.
For batches of at most 64 rows, \ksel uses 8 warps per thread block instead of 4; the batch size is fixed per CUDA graph,
so this choice is fixed as well. \katt uses 4 warps and 3 pipeline stages and processes one 64-token page per loop
iteration. We tried and discarded two variants: fusing \kcmb into \katt (the last thread block of each row merges all
partials) makes every row wait for its last-arriving thread block, and programmatic dependent launch between the kernels changed TPOT by
$+1.5\%$, $-1.2\%$ and $+1.4\%$ at 8k, 32k and 130k, with no clear benefit.

\section{Correctness Checks}
\label{app:correct}
This appendix expands the correctness paragraph of Section~\ref{sec:impl}.

We check three properties.
(1)~\emph{When every page is kept ($\gam{=}1$), \sys computes dense attention exactly.} Its output differs from FA3 by at
most one bf16 ulp ($3.9\times10^{-3}$), that is, one bf16 rounding step at this magnitude. CUDA-graph replay matches
eager execution bit for bit, and in mixed steps the prefill rows are bit-identical to FA3.
(2)~\emph{That difference is comparable to the difference already present in dense attention.} In greedy generation
of 256 tokens on 12 prompts, \sys at $\gam{=}1$ and dense produce identical outputs on 4/12 prompts, while dense at two
different batch sizes is identical on 5/12. Every first divergence falls on a token whose top-2 logits are only
0--0.25 apart, where rounding error alone can change the choice.
(3)~\emph{The kernels select the right pages.} The page lists of \ksel match a float32 reference implementation of the
rule on 64/64 test rows at $\gam{=}0.98$. All experiments in Section~\ref{sec:eval} use the same \kest and \katt kernels and the same coverage
path of \ksel.

\clearpage
\section{Kernel Time vs.\ Length Spread}
\label{app:cv}

Table~\ref{tab:cv} lists the data of Figure~\ref{fig:cv}.

\begin{table}[H]
\centering\small
\caption{\textbf{Attention time per layer (\textmu s) on the same page lists} (Llama-3.1-8B geometry; data of
Figure~\ref{fig:cv}). CV 0: all rows of equal length; CV 1: nominal spread (measured 0.84--0.88); real: page counts
from decode steps of \sys at about 8k tokens, applied as fractions of each context. FlashInfer needs its host plan at every step (plan: page counts copied to the
host, index arrays built and \texttt{plan()} called, per layer); ``--'': not measured. Bold: fastest
method per row, FlashInfer counted with its plan.}
\label{tab:cv}
\vskip 3pt
\setlength{\tabcolsep}{6pt}
\begin{tabular}{@{}lrlrrrrrrr@{}}
\toprule
& & & & \multicolumn{3}{c}{FA3} & \multicolumn{2}{c}{FlashInfer} & \\
\cmidrule(lr){5-7}\cmidrule(lr){8-9}
Shape & Rows & CV & \sys & varlen & pad & pow2 & kernel & plan & Dense \\
\midrule
\multirow{3}{*}{8k$\times$32} & \multirow{3}{*}{256} & 0 & 70.1 & \textbf{68.3} & 68.5 & 75.7 & 65.5 & 668 & \multirow{3}{*}{248} \\
 &  & 1 & \textbf{73.5} & 84.2 & 168.4 & 107.5 & 105.0 & 672 &  \\
 &  & real & \textbf{120.5} & 139.2 & 228.5 & 161.7 & 149.0 & 683 &  \\
\addlinespace[3pt]
\multirow{3}{*}{32k$\times$16} & \multirow{3}{*}{128} & 0 & 120.6 & \textbf{118.6} & 118.7 & 132.1 & 120.8 & 455 & \multirow{3}{*}{489} \\
 &  & 1 & \textbf{123.2} & 151.3 & 337.3 & 207.3 & 119.4 & 467 &  \\
 &  & real & \textbf{200.9} & 238.3 & 491.5 & 280.5 & 194.2 & 464 &  \\
\addlinespace[3pt]
\multirow{3}{*}{64k$\times$8} & \multirow{3}{*}{64} & 0 & \textbf{119.7} & 122.9 & 123.0 & 137.1 & -- & -- & \multirow{3}{*}{495} \\
 &  & 1 & \textbf{120.5} & 153.1 & 325.2 & 208.1 & -- & -- &  \\
 &  & real & \textbf{198.7} & 231.7 & 495.1 & 278.1 & -- & -- &  \\
\addlinespace[3pt]
\multirow{3}{*}{131k$\times$4} & \multirow{3}{*}{32} & 0 & \textbf{119.6} & 122.8 & 122.6 & 136.9 & 119.9 & 287 & \multirow{3}{*}{491} \\
 &  & 1 & \textbf{120.4} & 142.0 & 350.3 & 186.8 & 117.6 & 289 &  \\
 &  & real & \textbf{183.0} & 184.5 & 429.3 & 261.9 & 172.6 & 285 &  \\
\bottomrule
\end{tabular}
\end{table}

\paragraph{One page list shared by all KV heads.}
If all KV heads of a (request, layer) share one page list, the existing paged kernels of FA3 and FlashInfer run
unchanged, but the shared list must hold the union of the pages every head needs. On real page counts the shared list
keeps 78--87\% of the pages, against 33--41\% with one list per head, so applying the coverage rule at layer
granularity saves almost nothing. On the real shared list at 131k, \katt takes 396.1\,$\mu$s per layer, native FA3
424.9\,$\mu$s, and FlashInfer 387.1\,$\mu$s plus a 486.7\,$\mu$s host plan.

\paragraph{FlashInfer as the attention backend of \sys.}
On the page lists of \sys at 128k, FlashInfer's tensor-core page-1 kernel is faster than \katt (3.52 vs.\ 3.94\,ms per
step), but it needs a 2.16\,ms host plan. A device-side plan for that kernel would combine both advantages
(Section~\ref{sec:conclusion}).

\section{Additional Latency and Serving Results}
\label{app:extra}
This appendix complements Sections~\ref{sec:eval-tpot} and~\ref{sec:eval-serve}.

\paragraph{TPOT of adaptive vs.\ fixed budgets.}
We run $\gam{=}0.98$ and a fixed $K$ with the same fraction of page reads in vLLM. The coverage rule is 3.6\% slower at
8k$\times$32 (9.21 vs.\ 8.89\,ms), 5.5\% faster at 32k$\times$16 (8.50 vs.\ 8.99\,ms) and 6.4\% slower at
130k$\times$4 (7.63 vs.\ 7.17\,ms; across repetitions $\gam{=}0.98$ varies over 7.20--7.64\,ms). The differences have
no consistent sign, so a per-row budget has no consistent time cost beyond run-to-run variation. This comparison ran an earlier version of the fixed-count
mode of \ksel, which broke ties at the cut by page position. The fixed-$K$ and profiled rows of
Table~\ref{tab:ablation} and the fixed-count runs on RULER use the final version, which changes only how those ties are
resolved; both keep the same number of pages, so the timing is unaffected.

\paragraph{Mixed prefill and decode steps.}
Prefill rows of a mixed step run through unmodified FA3. On 64 LongBench requests (1.40M input tokens, 18{,}715 output
tokens) with continuous batching on Llama-3.1-8B, output throughput rises from 265.3--265.7 to 312.4--312.8\,tok/s
(+18\%) with $\gam{=}0.98$.

\paragraph{Serving details.}
Dense FlashInfer is no faster than dense FA3 (Llama-3.1-8B, set A, $c{=}16$: median TPOT 34.2 vs.\ 28.6\,ms), so the
gains of \sys hold against either dense backend. On set A, throughput at concurrency 32 is lower than at 16 for every
arm: 32 requests of up to 64k tokens do not fit in the KV cache (840k tokens for Llama-3.1-8B), the cache reaches
94--100\% usage while requests wait, and the batch is capped by memory. \sys saves bandwidth, not capacity, and is
capped in the same way.

\begin{table}[H]
\caption{\textbf{TPOT (ms) with $\gam{=}0.99$} and dense FA3, same shapes and protocol as Table~\ref{tab:tpot}
(mean of two interleaved pairs), and page reads of \sys. Bold: lower TPOT.}
\label{tab:tpot99}
\centering\small
\setlength{\tabcolsep}{2.2pt}
\begin{tabular}{@{}lrrrrrrrr@{}}
\toprule
& \multicolumn{4}{c}{Llama-3.1-8B} & \multicolumn{4}{c}{Qwen3-4B} \\
\cmidrule(lr){2-5}\cmidrule(l){6-9}
Shape & dense & \sys & $\times$ & reads & dense & \sys & $\times$ & reads \\
\midrule
8k$\times$32   & 12.84 & \textbf{10.29} & 1.25 & 0.55 & 12.37 & \textbf{10.02} & 1.23 & 0.60 \\
16k$\times$32  & 19.57 & \textbf{13.10} & 1.49 & 0.49 & 20.76 & \textbf{14.00} & 1.48 & 0.54 \\
32k$\times$16  & 16.65 & \textbf{9.91}  & 1.68 & 0.43 & 17.15 & \textbf{10.73} & 1.60 & 0.50 \\
64k$\times$8   & 19.70 & \textbf{9.12}  & 2.16 & 0.24 & 20.66 & \textbf{10.22} & 2.02 & 0.34 \\
130k$\times$4  & 19.50 & \textbf{8.59}  & 2.27 & 0.22 & 20.49 & \textbf{9.03}  & 2.27 & 0.28 \\
\bottomrule
\end{tabular}
\end{table}

\begin{figure}[H]
  \centering
  \includegraphics[width=\textwidth]{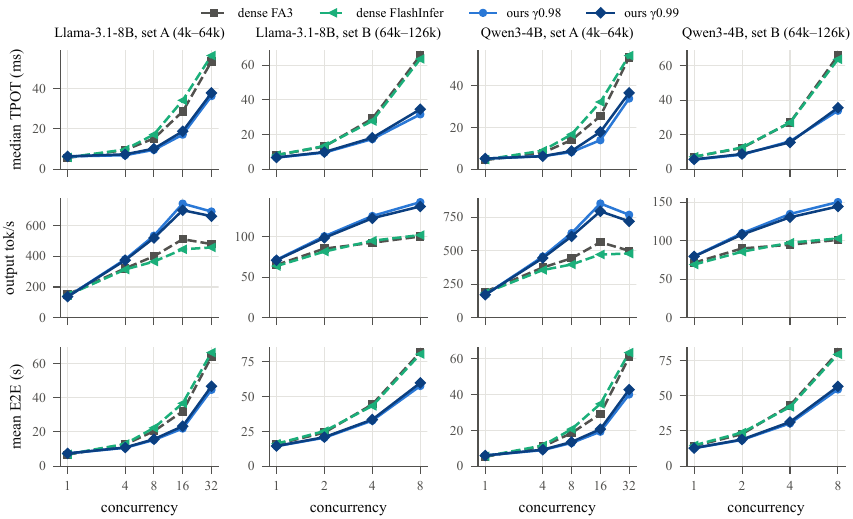}
  \vspace{-6mm}
  \caption{\textbf{Online serving with \texttt{vllm bench serve}} (closed loop, prefix caching off, 1{,}024 output
  tokens, one run per point). Set A: LongBench prompts of 4k--64k tokens; set B: 64k--126k tokens.}
  \label{fig:serve}
\end{figure}

\begin{figure}[H]
  \centering
  \includegraphics[width=0.6\textwidth]{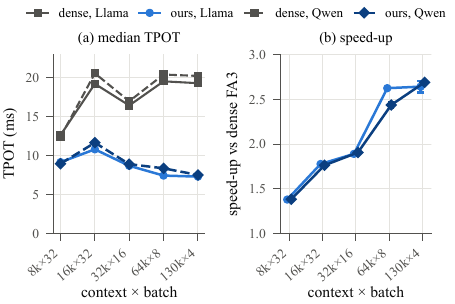}
  \vspace{-6mm}
  \caption{\textbf{TPOT in vLLM} (\sys at $\gam{=}0.98$ vs.\ dense FA3; data of Table~\ref{tab:tpot}). Median of 5
  repetitions, two interleaved pairs per shape; the error bars span the two pairs.}
  \label{fig:tpot}
\end{figure}

\section{Breakdown of the Comparison with Prior Methods}
\label{app:comp}
\begin{figure}[H]
  \centering
  \includegraphics[width=\textwidth]{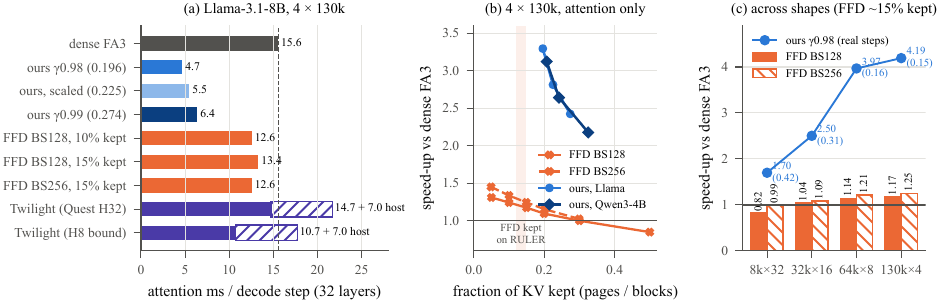}
  \vspace{-6mm}
  \caption{\textbf{Sparse decode methods vs.\ dense FA3, attention only} (Section~\ref{sec:eval-comp}). (a)~One decode
  step (32 layers) of Llama-3.1-8B at 4$\times$130k. Twilight: the authors' selection chain and FlashInfer attention on
  its own token sets (GPU), plus host metadata and plan (hatched). (b)~Speedup against the kept fraction at the same
  geometry (\sys: 64-token pages; FFD: 128/256-token blocks at a preset kept fraction; shaded band: FFD's kept fraction
  on RULER). \sys uses page counts from a real decode step, scaled to the reads of $\gam{=}0.98$ and $0.99$ on RULER and,
labelled ``scaled'', to an intermediate fraction. (c)~Across shapes: FFD at about 15\% kept; \sys on real $\gam{=}0.98$ steps (kept fraction in
  parentheses).}
  \label{fig:comp}
\end{figure}

This appendix expands Section~\ref{sec:eval-comp}; Figure~\ref{fig:comp} summarizes the timings.

\paragraph{Protocol.} The Attn and Host columns of Table~\ref{tab:main} time one decode step at
4$\times$130{,}048 tokens on the same GPU. Each prior method runs its authors' kernels where they exist, with the
token-set size it keeps on RULER. For \sys, the page count of each (request, layer, KV head) is taken from real decode
steps and scaled to the reads measured on RULER. TPOT is measured end to end with CUDA graphs on eight RULER 128k
prompts per model (two batches of four prompts, 200 decode steps, median of five runs per batch). Dense, \sys,
FA3\textsuperscript{c} and FlashInfer\textsuperscript{c} run in vLLM; the last two replace only the attention call of
\sys and read the same page lists. FA3 treats each (request, KV head) as a varlen sequence over vLLM's head-major (HND) cache layout and
builds its split metadata in every layer. FlashInfer is planned once per batch shape with every row at full width,
and at every step device code writes the real index arrays into the planned buffers. FFD runs in its own runtime
(Hugging Face with full CUDA graphs, Llama only), on the same prompts cut to equal length (first 256 and last tokens),
because its cache holds one length per batch. Quest and Twilight have no vLLM path and are compared by Attn and Host.

\paragraph{Quest: slow because of the kernel, page-hungry because of the rule.}
Quest differs from \sys in two places. The first is the kernel: Quest's rule selects pages per query head and
attends with FlashInfer's paged kernel (CUDA-core kernel in both arms below). On our page sets at 128k, FlashInfer with one list per query head takes 8.30\,ms, against
4.89\,ms with one list per KV head, before a 2.16\,ms host plan. The kernel of \sys removes this cost for any rule,
Quest's bound included (Section~\ref{sec:generality}).
The second is the rule. On Llama-3.1-8B, $\gam{=}0.98$ matches the accuracy of Quest-36k with only 19.6\% of the pages,
while Quest's budget corresponds to about 27.4\%. On our kernel these two read levels cost 4.73 and 6.43\,ms per step,
so the rule alone makes attention \textbf{1.36$\times$} faster at equal accuracy. With both differences combined,
Quest's own path takes 22.0\,ms, 1.4$\times$ dense attention (Table~\ref{tab:main}).

\paragraph{FFD: slow because it scans every key.}
FFD keeps 12.5\% (Llama-3.1-8B) and 21.2\% (Qwen3-4B) of the 128-token blocks on RULER, yet its fused kernel is only
1.20$\times$ and 1.08$\times$ faster than dense, and on Llama-3.1-8B its measured TPOT equals dense (18.8 vs.\
19.0\,ms). The reason is that at every step it scans the 2-bit copy of all keys,
so its cost barely falls with the number of kept blocks: even at 2\% kept it is only 1.36$\times$ faster, and at
8k$\times$32 it is slower than dense at every kept fraction (Figure~\ref{fig:comp}). Across batch
shapes, FFD runs at 0.82--1.25$\times$ the speed of FA3, whereas the attention path of \sys is
\textbf{1.70--4.19$\times$} faster than FA3 (Table~\ref{tab:nsys}).

\paragraph{Twilight: slow because it selects per query head.}
Twilight's selection chain takes 12.71\,ms per step, because its kernels treat GQA as 32 independent heads (8.72\,ms
with an 8-head port). Attention adds 2.02\,ms, and the host plan takes 6.96\,ms outside any CUDA graph. Selection and
attention alone take 14.73\,ms, close to dense attention (15.59\,ms); with the host plan, 21.69\,ms, against
\textbf{4.73\,ms} for \sys at $\gam{=}0.98$.

\end{document}